%% file: main.tex
\documentclass[lettersize,10 pt, journal, twoside]{IEEEtran}
\usepackage{amsmath,amsfonts}
\usepackage{array}
\usepackage[caption=false,font=footnotesize]{subfig}
\usepackage{textcomp}
\usepackage{stfloats}
\usepackage{url}
\usepackage{verbatim}
\usepackage{graphicx}
\usepackage{cite}

\usepackage{dsfont}   
\usepackage{pifont}   
\usepackage{gensymb}   
\usepackage{textgreek} 
\usepackage{xcolor}
\usepackage{tikz} 
\usepackage[caption=false,font=footnotesize]{subfig}
\usepackage[linesnumbered,ruled,vlined]{algorithm2e}
\usepackage[pdfencoding=auto, hidelinks]{hyperref} 

\newcommand{\refeqn}[1]{Eq.~\ref{#1}}
\newcommand{\reffig}[1]{Fig.~\ref{#1}}
\newcommand{\refsec}[1]{Sec.~\ref{#1}}
\newcommand{\refalg}[1]{Algorithm~\ref{#1}}
\newcommand{\reftab}[1]{Table~\ref{#1}}

\newcommand{\footnoteurl}[2]{\footnote{\tt\small\href{#1}{#2}}}
\newcommand{\xmark}{\ding{51}}  

\newcommand{\norm}[1]{\left\lVert#1\right\rVert}
\DeclareMathOperator{\std}{std}

\DeclareMathOperator*{\argmaxD}{argmax}

\definecolor{forestgreen}{rgb}{0.13, 0.55, 0.13}
\definecolor{firebrick}{rgb}{0.7, 0.13, 0.13}

\begin{document}

\title{End-to-End Optimization of LiDAR Beam Configuration for 3D Object Detection and Localization}

\author{
Niclas Vödisch$^{1,2}$,
Ozan Unal$^{1}$,
Ke Li$^{1}$,
Luc Van Gool$^{1,3}$,
and Dengxin Dai$^{4}$
\thanks{© 2022 IEEE. Personal use of this material is permitted. Permission from IEEE must be obtained for all other uses, in any current or future media, including reprinting/republishing this material for advertising or promotional purposes, creating new collective works, for resale or redistribution to servers or lists, or reuse of any copyrighted component of this work in other works.}
\thanks{$^{1}$ Niclas Vödisch, Ozan Unal, Ke Li, and Luc Van Gool are with the Computer Vision Lab, ETH Zürich, Switzerland.}%
\thanks{$^{2}$ Niclas Vödisch is also with Autonomous Intelligent Systems, University of Freiburg, Germany. {\tt\footnotesize voedisch@cs.uni-freiburg.de}}%
\thanks{$^{3}$ Luc Van Gool is also with Processing Speech and Images, KU Leuven, Belgium.}%
\thanks{$^{4}$ Dengxin Dai is with Vision for Autonomous Systems, MPI for Informatics, Saarbrücken, Germany.}%
\thanks{Digital Object Identifier 10.1109/LRA.2022.3142738}%
}

\markboth{T\MakeLowercase{his paper appeared in:} IEEE ROBOTICS AND AUTOMATION LETTERS, VOL. 7, ISSUE 2, APRIL 2022}%
{Vödisch \MakeLowercase{\textit{et al.}}: End-To-End Optimization of LiDAR Beam Configuration for 3D Object Detection and Localization}


\maketitle


\begin{abstract}
Existing learning methods for LiDAR-based applications use 3D points scanned under a pre-determined beam configuration, e.g., the elevation angles of beams are often evenly distributed. Those fixed configurations are task-agnostic, so simply using them can lead to sub-optimal performance. In this work, we take a new route to learn to optimize the LiDAR beam configuration for a given application. Specifically, we propose a reinforcement learning-based learning-to-optimize (RL-L2O) framework to automatically optimize the beam configuration in an end-to-end manner for different LiDAR-based applications. The optimization is guided by the final performance of the target task and thus our method can be integrated easily with any LiDAR-based application as a simple drop-in module. The method is especially useful when a low-resolution (low-cost) LiDAR is needed, for instance, for system deployment at a massive scale. We use our method to search for the beam configuration of a low-resolution LiDAR for two important tasks: 3D object detection and localization. Experiments show that the proposed RL-L2O method improves the performance in both tasks significantly compared to the baseline methods. We believe that a combination of our method with the recent advances of programmable LiDARs can start a new research direction for LiDAR-based active perception. The code is publicly available at \href{https://github.com/vniclas/lidar_beam_selection}{\texttt{github.com/vniclas/lidar\_beam\_selection}}.
\end{abstract}




\input{sections/1_introduction}
\input{sections/2_related_work}
\input{sections/3_method}

\input{sections/4_experiments}
\input{sections/5_conclusion}


\bibliography{references}{}
\bibliographystyle{IEEEtran}


\vfill
\end{document}

%% file: sections/1_introduction.tex
\section{Introduction}
\label{sec:introduction}

\IEEEPARstart{A}{ctive} depth measurement systems, particularly LiDARs, are key sensors in the design of current autonomous vehicles. Due to their ability to precisely capture 3D scenes, they have been applied to a variety of robotics problems, e.g., object detection~\cite{Meng_2020_ECCV}, depth completion~\cite{Bergman_2020_ICCP}, mapping~\cite{Yang_2018_IROS}, and localization~\cite{Elbaz_2017_CVPR}. Since most of these tasks benefit from a higher resolution, i.e., more measurement points, there has been an incentive in the LiDAR industry to continuously increment the number of beams of mechanical rotating scanners, currently reaching up to 128 lasers, e.g., the Velodyne Alpha Prime and RoboSense RS-Ruby. 

However, such high-resolution~(HR) LiDARs come at a high cost -- an HR LiDAR with 128 lasers can cost as much as USD 60,000. Therefore, there is a strong need to develop methods that work well with low-resolution~(LR) LiDARs. In response to this, numerous methods have been proposed to improve the performance of autonomous applications with LR LiDAR data \cite{Pino_2018_ROBOT,Liao_2017_ICRA,Lin_2020_ICRA,Bai_2021_Arxiv}. While this stream of research is interesting, the beam configurations of the used LR LiDARs are fixed (often pre-determined by LiDAR manufacturers), i.e., the beam configurations are task-agnostic and not optimal for the tasks at hand. LiDAR beams at different positions can contribute differently to the performance of different tasks. For instance, for 3D object detection, beams directing to areas where most objects appear are more useful than others. For localization, beams scanning over static parts of the scene are more important. This highlights the fact that the setting used by those existing methods -- improving data processing only while keeping the LiDAR beam configurations fixed -- is not optimal.

\begin{figure}
    \centering
    \includegraphics[width=\linewidth]{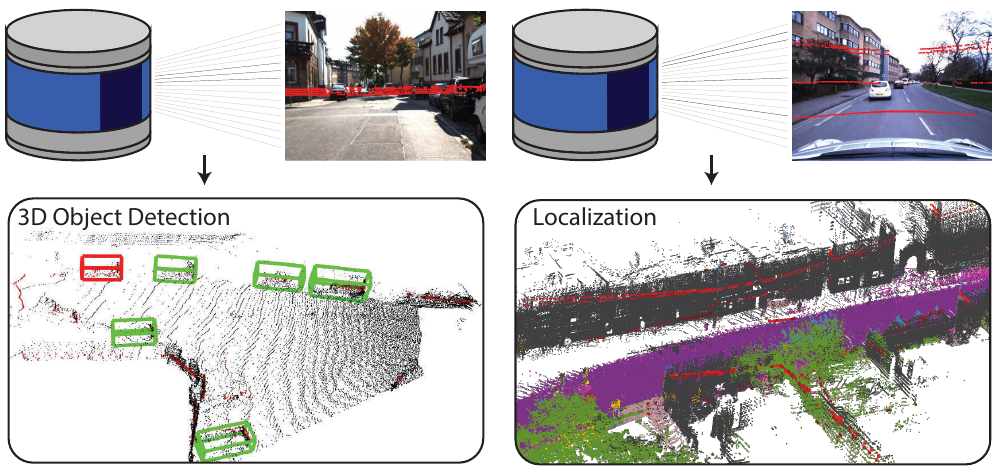}
    \caption{Illustration of end-to-end optimization of LiDAR beam configuration for 3D object detection and localization. The 4-beam solution space is simulated via the sampling of beams from a high-resolution LiDAR. The vast search space is then efficiently traversed to find a high-performing configuration for each individual task.}
    \label{fig:overview}
\end{figure}

This leads to the following question: \emph{Can the best beam configuration of LiDARs be learned for a given target task?} This work shows that this is possible and doing so can lead to pronounced performance improvements for different LiDAR-based applications. In \reffig{fig:overview}, we demonstrate the optimization results of four LiDAR beams for 3D object detection and localization. The importance of this work is supported by the recent advances in LiDAR technologies~\cite{Pittaluga_2020_3DV, Nakamura_2021_OPTO}, showing that adaptive sampling and dynamically reprogramming LiDAR beam patterns is possible. A marriage of our method with theirs will enable utilizing different configurations for different tasks and scenes, which will open a new avenue for active perception. 

To avoid a multi-stage method, we propose a reinforcement learning-based learning-to-optimize (RL-L2O) framework to automatically optimize the beam configuration in an end-to-end manner for different LiDAR-based applications. The method is automated from two aspects: First, it automates the exploration of the solution space, avoiding the difficulty of manually proposing potential candidates. Second, it uses the final performance of the downstream application as the optimization guidance, such that the method can be easily integrated with any LiDAR-based application as a simple drop-in module without any changes to the design. 

Overall, this work makes the following contributions: 
\begin{itemize}
    \item We provide a novel perspective to improve the performance of LiDAR-based applications, by learning to position LiDAR beams to the most important positions for a given task. This is especially useful when a low-resolution, i.e., low-cost, LiDAR is preferred.
    \item We propose a novel learning method for this new problem based on reinforcement learning. The method is end-to-end trainable and can be integrated with any LiDAR-based application as a simple drop-in module.
    \item We showcase the capability of our method on two of the most important LiDAR-based applications: 3D object detection and localization. Extensive experiments prove the efficacy of our method -- optimizing the beam configuration with our method can significantly boost the performance over the results of baseline methods.
\end{itemize}

%% file: sections/2_related_work.tex
\section{Related Work}
\label{sec:related_work}

\noindent \textbf{Adaptive LiDAR:}
Recent improvements in LiDAR technology, e.g., MEMS omnidirectional scanning~\cite{Wang_2019_Transducers}, as well as proof-of-concepts setups for both MEMS-based LiDARs~\cite{Pittaluga_2020_3DV} and optical phased arrays~\cite{Nakamura_2021_OPTO} give rise to software-side adaptive sampling. There are three depth estimation methods that share a similar spirit to our work. Specifically, Pittaluga et al.~\cite{Pittaluga_2020_3DV} demonstrate the flexibility of their laboratory LiDAR by adapting the scan pattern based on RGB images. Both Berman et al.~\cite{Bergman_2020_ICCP} and Gofer et al.~\cite{Gofer_2021_IP} learn to sub-sample depth measurement points from an HR LiDAR to enhance monocular depth estimation. The selected points can come from any beam of the HR LiDAR, i.e., all beams of the HR LiDAR are still needed. This raises a grand challenge to deliver it as a low-cost system. Our method selects LiDAR beams directly and is much easier to deploy.

\noindent 
\textbf{Hyperparameter Optimization:}
The goal of hyperparameter optimization (HPO) is to select parameters of a model from a configuration space that maximize the model's performance. HPO methods can be mainly classified into two groups: 1) naive approaches traverse the search space without any guidance, e.g., grid search; 2) sample-based methods select parameters based on prior data to optimize for a policy, e.g., Bayesian optimization~\cite{Wu_2019_EST}. Since evaluations of a model can be expensive, methods such as sequential model-based global optimization (SMBO)~\cite{Koide_2021_ICRA} introduce a surrogate that approximates the true model while being cheap to evaluate. Then, the true model is only evaluated for the parameters optimizing the surrogate.

Reinforcement learning (RL) poses yet another approach to guide HPO. The basic concept of RL is teaching an agent how to map a state to an action in order to maximize a received reward signal~\cite{Sutton_2018_book}. Wu et al.~\cite{Wu_2020_NC} model HPO as a Markov decision process to sequentially select hyperparameters and train a neural network to predict the performance of the current selection. Dong et al.~\cite{Dong_2021_TPAMI} employ Q-learning to optimize the network hyperparameters of an existing object tracker.
Similarly, we train a small neural network to estimate the performance of a downstream LiDAR task given a proposed beam configuration.

\noindent \textbf{3D Object Detection using Sparse LiDAR Data:}
3D object detection is the task of simultaneous recognition and localization of objects such as cars and pedestrians. Current state-of-the-art detectors solely rely on LiDAR point clouds as input due to their inherent accurate depth information~\cite{Lang_2019_CVPR, Unal_2021_WACV, Shi_2019_CVPR, Deng_2021_AAAI, Yan_2018_Sensors}. Models that only work on RGB camera images show drastically lower precision when estimating bounding boxes in 3D space. However, unlike monocular and stereo camera systems, the LiDAR sensor still remains an infeasible option in deployment due to its accompanying high manufacturing costs.

To this end, Wang et al.~\cite{Wang_2019_CVPR} propose generating pseudo LiDAR point clouds using RGB camera information by estimating depth maps. This allows existing 3D pipelines to be used to reduce the gap between the two domains, resulting in a more precise estimation of 3D bounding boxes while utilizing only RGB information. Pseudo-LiDAR++~\cite{Yurong_2020_ICLR} extends this idea by introducing an additional inexpensive sparse LiDAR sensor with only four beams to graphically correct the estimated depth maps, which in turn increases the detection performance for all models. While Pseudo-LiDAR++ shows great improvements in 3D object detection by incorporating equidistant 4-beam LiDAR information, the beam selection process remains unexplored. In this work, we show that our proposed method can be used to learn a better and robust sparse LiDAR configuration, which outperforms an arbitrarily chosen equidistant beam selection.

\noindent \textbf{LiDAR-based Mapping and Localization:}
Constructing 3D maps from LiDAR point clouds on a city-scale level allows for later localization using 3D registration. Unlike in online LiDAR-SLAM systems, e.g., LOAM~\cite{Zhang_2017_AR} and its variants, here the goal is to obtain highly accurate point clouds by exploiting offline optimization. An often followed approach is to divide large areas into smaller submaps and thereby decompose the problem into two separate optimization steps, i.e., intra- and inter-loop optimization~\cite{Shiratori_2015_3DV}. Commonly, the mapping process is implemented using a pose graph~\cite{Mendes_2016_SSRR} to jointly optimize ICP-based LiDAR odometry and INS data~\cite{Yang_2018_IROS}. LiDAR-based localization is frequently solved by direct 3D registration, e.g., using ICP~\cite{Chen_1992_IVC} as the main method~\cite{Mendes_2016_SSRR}. Alternatively, precise 3D registration can be exploited to fine-tune the results of a previous coarse localization step~\cite{Elbaz_2017_CVPR}. Although the complexity of ICP is a function of the number of points, there has been little work done on optimal point sampling from an HR LiDAR while maintaining reliable localization results. In this work, we follow the popular pose graph-based approach for our map generation and then optimize the design of an LR LiDAR targeting robust localization with sparse data.

%% file: sections/3_method.tex
\section{Method}
\label{sec:method}

\begin{figure*}[t]
    \centering
    \includegraphics[width=\textwidth]{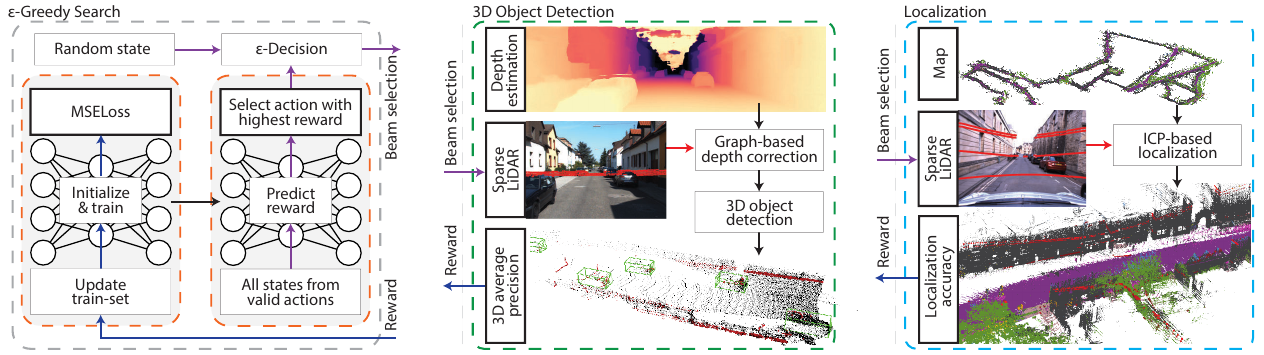}
    \caption{Illustration of our proposed reinforcement learning-based  learning-to-optimize (RL-L2O) framework with two possible reward pipelines for 3D object detection and localization. (left) Block diagram of the \textepsilon-GS algorithm. In every iteration, the agent is trained using an online training set consisting of pairs of beam sets and corresponding rewards. The agent then predicts a reward for all actions and selects the action with the highest expected reward. Based on an \textepsilon-probability, the agent either applies its chosen action or explores a random state. (middle) Overview of the 3D detection pipeline used to generate a mapping from a set of LiDAR beams to a reward in $\mathbb{R}$ via the 3D mean AP value of a trained 3D object detector. (right) Reward design for sparse LiDAR-based localization by comparing results from ICP with ground truth poses.}
    \label{fig:system_overview}
\end{figure*}

In this section, we introduce our reinforcement learning-based learning-to-optimize (RL-L2O) method to automatically search for the optimal beam configuration for a given LiDAR-based task. In our RL-L2O framework, we aim to learn a parameterized policy $\pi$ to dynamically improve the current beam configuration given the previous experience. The goal of RL-L2O is to learn a policy $\pi^*$ that maximizes the overall expected reward $r$ over time: 
\begin{equation}
    \pi^* = \argmaxD_\pi \mathbb{E}_{\mathbf{s}_0,\mathbf{a}_0,\mathbf{s}_1,...,\mathbf{s}_T} \left[ \sum_{t=0}^T r_t \right]
\end{equation}
where $T$ is the length of the sequence of states and actions. An action $\mathbf{a}_t$ is used to update the beam configuration $\mathbf{s}_t$ that we call a state. The reward $r_t$ signals the performance of the optimization, which is defined by the final performance of the given task.

\subsection{Design of Solution Space}
Let $k$ be the total number of LiDAR beams that we would like to have for the LR LiDAR and let $[\varphi_l, \varphi_h]$ be the  range of elevation angles for the laser beams. If beams could be positioned at any place within this range, the solution space and action space would be too large. To avoid this difficulty, we limit the candidate positions to the $K (K>k)$ beam positions of an HR LiDAR. This way, we can use the data recorded by an HR LiDAR to simulate the data of the optimized LR LiDAR.  
Given a task and the dataset recorded by a specific HR LiDAR, $\varphi_l$, $\varphi_h$, and $K$ are determined. We thus can just use beam IDs $\mathbf{s}_t=(s_t^1, s_t^2, \dots, s_t^k)$ to represent the selected beam locations, where $s_t^i \in [1, K]$ $\subset \mathbb{N}$ for all $i \in [1, k]$. Note that the beam IDs are sorted in a strictly ascending order.

\subsection{Design of the Action Space}
Our action vector $\mathbf{a}_t$ has the same shape as $\mathbf{s}_t$, i.e., $\mathbf{a}_t=(a^1_t, a_t^2, \dots, a_t^k)$ with $a_t^i \in [-m,m] \subset \mathbb{Z}$. $m$ is a small value used to specify the strength of the action. The way an action~$\mathbf{a}_t$ changes a state~$\mathbf{s}_t$ is defined as: 
\begin{equation}
    \mathbf{s}_{t+1} = \mathbf{s}_t + \mathbf{a}_t = (s^1_t + a^1_t, s^2_t + a^2_t, \dots, s^k_t + a^k_t). 
\end{equation}
\noindent
We further restrict the action space to those actions that yield states withing the solution space, i.e., $\mathbf{a}_t$ is valid if $\mathbf{s}_{t+1}^i \in [1, K]$ for all $i \in [1, k]$ and the beam IDs of $\mathbf{s}_{t+1}$ are unique.

\subsection{Value Function Approximation}
\label{sec:method_reward_prediction}
We use a neural network to estimate the state-value function. In every iteration, the network is re-initialized and trained from scratch to incorporate the latest state-value pair. The network takes as input a feature vector that summarizes all characteristics of the beam configurations to predict the value $\hat{v}(\mathbf{s})$ for a given state $\mathbf{s}$. 

We design the input feature $\mathbf{f} \in \mathbb{R}^h$ to be both descriptive and compact. The feature space consists of two types of features. First, beam-wise features are generated individually for each beam ID. Second, pairwise features are computed using two beams and hence reflect the relationship among them. In total, we propose five individual and one pairwise features: number of points $f_{pts}$, number of points of each semantic class $f_{sem\_pts}^{c'}$, mean distance $f_{dist}$, standard deviation of the mean distance $f_{std\_dist}$, elevation angle $f_{\varphi}$, and the pairwise elevation angle difference $f_{\varphi\_diff}$.
Let $b_s$ be a LiDAR beam of ID $s$ with $N$ points $p^l=(x^l, y^l, z^l, c^l)$ for all $l \in [1, N]$, where $(x, y, z) \in \mathbb{R}^3$ are the 3D Cartesian coordinates and $c \in C$ is the semantic label (details in \refsec{sec:implementation_details}).
Additionally, let $\{b_s^i | i = 1, 2, \dots, M\}$ be a set of $M$ LiDAR scans taken by beam $s$.
We compute the features by averaging over $M$ scans and, if applicable, over all points in a beam.
The individual features are defined as follows\footnote{To improve readability, we omit the beam ID from the notations.}:

\vspace{-1ex}

\begin{equation}
\begin{split}
    f_{pts} &= \tfrac{1}{M} \textstyle \sum_{i=1}^{M} N_i \\
    f_{sem\_pts}^{c'} &= \tfrac{1}{M} \textstyle \sum_{i=1}^{M} \textstyle \sum_{l=1}^{N_i} \mathds{1}_{c'}\left( c_i^l \right), \forall c' \in C \\
    f_{dist} &= \tfrac{1}{M} \textstyle \sum_{i=1}^{M} \tfrac{1}{N_i} \textstyle \sum_{l=1}^{N_i} \norm{\left(x_i^l, y_i^l\right)} \\
    f_{std\_dist} &= \tfrac{1}{M} \textstyle \sum_{i=1}^{M} \std \left(\norm{\left(x_i^l, y_i^l\right)}\right), l = 1, 2, \dots, N_i \\
    f_{\varphi} &= \tfrac{1}{M} \textstyle \sum_{i=1}^{M} \tfrac{1}{N_i} \textstyle \sum_{l=1}^{N_i} \arcsin \left( \tfrac{z_i^l}{\norm{\left(x_i^l, y_i^l\right)}} \right)
\end{split}
\end{equation}

\noindent
The pairwise feature between two beams $s_i$ and $s_j$ with \mbox{$s_i < s_j$} is defined as:

\vspace{-1ex}

\begin{equation}
    f_{\varphi\_diff} = f_{\varphi, s_j} - f_{\varphi, s_i}
\end{equation}

\noindent
Consequently, the complete feature vector~$\mathbf{f}$ of a state~$\mathbf{s}$ is:
\begin{equation}
\begin{aligned}
    \mathbf{f} ={} & [f_{s_i, pts}, f_{s_i, sem\_pts}^{c'}, f_{s_i, dist}, f_{s_i, std\_dist}, f_{s_i, \varphi}, \dots \\
          & \phantom{[} f_{s_i, s_j, \varphi\_diff}, \dots ]
\end{aligned}
\label{eqn:feature_vector}
\end{equation}
with $j = i + 1$, $\forall i, j \in [1,k]$ and $\forall c' \in C$.

The real value, i.e., the ground truth value, $v(\mathbf{s})$ of state $\mathbf{s}$ is computed by actually training and evaluating a model for the target task with data obtained from beam configuration~$\mathbf{s}$. The step is further detailed in \refsec{sec:method_object_detection} and \refsec{sec:method_lidar_based_localization} for 3D object detection and localization, respectively. 

\subsubsection{3D Object Detection}
\label{sec:method_object_detection}

We follow the standard pipeline introduced in~\cite{Yurong_2020_ICLR} for 3D object detection from stereo camera images and sparse LiDAR information that can be summarized in four steps: (1)~by utilizing stereo RGB camera information,  per-image depth maps are estimated; (2)~using $k$-beam LiDAR information, graph-based depth correction (GDC) is applied to reduce the initial error in the depth estimation; (3)~the corrected depth maps are used with corresponding pixel coordinates to generate pseudo LiDAR point clouds; and (4) the resulting point clouds are used to train an off-the-shelf 3D object detector.

\begin{algorithm}[t]
    \SetAlgoLined
    \SetAlgoNoEnd
    $train\_set \leftarrow$ SampleTrainSet($initial\_size$)\\
    $predictor \leftarrow$ TrainPredictor($train\_set$)\\
    $history \leftarrow$ InitializeHistory($train\_set$)\\
    $state \leftarrow$ PickBestState($history$)\\
    \For{t = initial\_size, ..., T}{
        \eIf{\upshape{SampleNumber(min=0, max=1)} $< \varepsilon$}{
            $action \leftarrow$ SampleRandomAction()
        }{
            $action \leftarrow$ GetBestAction($state, predictor$)
        }
        $state \leftarrow state + action$\\
        
        $history \leftarrow$ AddToHistory($state$)\\

        $value \leftarrow$ ComputeValue($state$)\\
        \If{$state$ \upshape{not in} $train\_set$}{
            $train\_set \leftarrow$ AddToTrainSet($state, value$)\\
            $predictor \leftarrow$ TrainPredictor($train\_set$)
        }
    }
    \textbf{return} PickBestState($history$)
    \BlankLine
    \SetKwProg{Fn}{function}{:}{}
    \Fn{\upshape{GetBestAction(}$state, predictor$\upshape{)}}{
        \ForEach{action}{
            $\widehat{state} \leftarrow state + action$\\
            $\widehat{value} \leftarrow$ PredictValue($predictor$, $\widehat{state}$)
        }
    }
    \textbf{return} $action$ leading to $\widehat{state}$ with largest $\widehat{value}$
    \caption{\textepsilon -Greedy Search}
    \label{alg:eps_greedy}
\end{algorithm}

Because the estimation of depth maps relies solely on stereo image information in step (1), the results are highly prone to errors. Therefore, the beam selection for GDC in step (2) becomes crucial for high performance, as the closer a pixel is to a projected LiDAR measurement the less bias it will carry over from the initial estimation. We employ the proposed RL-L2O to find a high-performing beam configuration, via an agent that learns a mapping between any set of beams and the respective performance when used in the standard detection pipeline. In other words, by fixing the model architecture and the initial depth estimation, the 3D object detection precision becomes a function of the beam configuration used in the depth correction step as illustrated in \reffig{fig:system_overview}~(middle). We base our reward directly on the most commonly used evaluation metric for 3D object detection, which is the 3D mAP (IoU $\geq 0.7$) of the moderate difficulty car class:
\begin{equation}
    \label{eq:reward_obj}
    v(\mathbf{s})_{obj} = mAP_{car}^{moderate}
\end{equation}

\subsubsection{LiDAR-Based Localization}
\label{sec:method_lidar_based_localization}

We perform LiDAR-based localization with respect to a previously created 3D map and compare the computed pose to a ground truth reference pose generated under human supervision. To create the required high-definition 3D map from LiDAR measurements, we leverage graph optimization following a three-step procedure: (1) pre-processing of the poses of the mapping vehicle; (2) construction and optimization of the resulting pose graph; and (3) accumulation of a globally consistent point cloud without movable objects.

During the execution of RL-L2O, a set of single LiDAR scans is used to localize within the generated 3D map as illustrated in Fig.~\ref{fig:system_overview}~(right). In particular, we first perform coarse registration using the GNSS position, followed by a fine-tuning step with point-to-plane ICP~\cite{Chen_1992_IVC}.

To guide the agent, we compute the reward as a weighted average of three localization accuracies for different matching thresholds proposed in~\cite{Sattler_2018_CVPR}: \mbox{(0.25m, 2\degree)}, \mbox{(0.50m, 5\degree)}, \mbox{(5.00m, 10\degree)}, referred to as $acc_1$, $acc_2$, and $acc_3$ respectively.
\begin{equation}
    v(\mathbf{s})_{loc} = \lambda_1 \, acc_1 + \lambda_2 \, acc_2 + \lambda_3 \, acc_3
    \label{eqn:reward_loc}
\end{equation}
with $\lambda_1$, $\lambda_2$, and $\lambda_3$ denoting the corresponding weights.

\subsection{\texorpdfstring{$\varepsilon$}{e}-Greedy Search}
\label{sec:method_beam_selection}
Since the agent implementation is decoupled from the environment, the \textepsilon-Greedy Search (\textepsilon -GS) can easily be applied. We outline the method in \refalg{alg:eps_greedy}. As opposed to a randomized traversal of the search space, the main idea of \textepsilon -GS is to learn a function that describes the relation between a task's performance and any set of beam IDs and to take the best actions accordingly. The learning is guided by a reward $r_t=v_t-v_{t-1}$. 
To warm start the value predictor, a small set of beam configurations is randomly chosen and the corresponding true value is computed by the environment. Furthermore, the best performing beam configuration out of the initial samples is chosen as the initial state $\mathbf{s}_0$. Until a predetermined number of environment requests $T$ is reached, the following steps (1) to (4) are performed as illustrated in \reffig{fig:system_overview}~(left): (1) Given all state-value pairs, a neural network (value predictor) is trained to learn the mapping $v_{task}(\mathbf{s})$; (2) With a probability $\varepsilon$ a random action is sampled. Otherwise, the value predictor is used to select the action with the highest expected reward given the current state $\mathbf{s}_t$. This allows balancing between exploration and exploitation; (3) The state $\mathbf{s}_t$ is updated by applying the chosen action; (4) If the new state $\mathbf{s}_{t+1}$ has not been visited in a previous iteration, the true value is computed by the environment and the state-value pair is added to the training set; (5) Finally, the best-performing state based on the true value is selected.

%% file: sections/4_experiments.tex
\begin{figure*}[t]
    \centering
    \includegraphics[width=\textwidth]{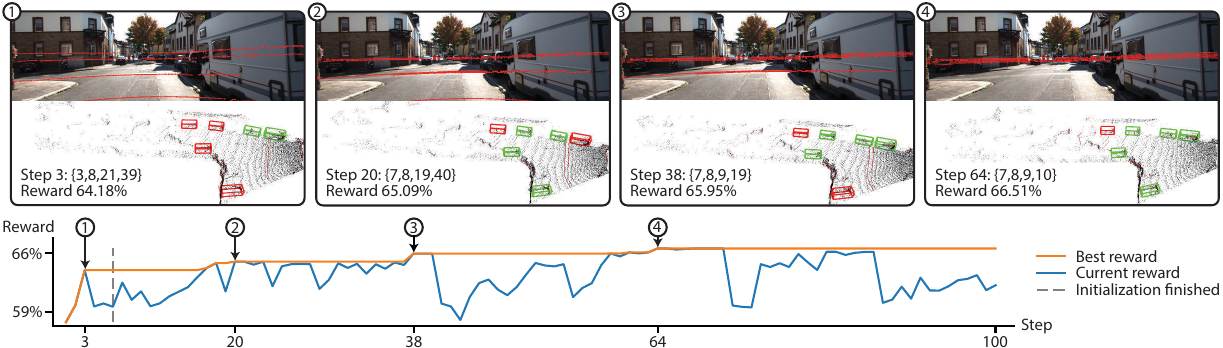}
    \caption{Evolution of the beam selection algorithm for 3D object detection. (bottom) For the plot, please refer to the legend on the right. (top) Four cells are presented to visualize the best-performing beam configuration at different stages of the algorithm. The resulting LiDAR beams are shown on both the camera image and the respective point cloud in red. As can be seen, the algorithm converges slowly to a state where all beams lie near the horizon to cover most (moderate difficulty) car points. The respective 3D object detection results are shown for each highlighted step. The bounding boxes are colored green if the estimation is a true positive and red otherwise.}
    \label{fig:search_evolution_obj}
\end{figure*}%

\begin{table*}
    \centering
    \caption{Evaluation of the 3D object detection performance (IoU $\geq 0.7$, mAP [\%]) on the KITTI~\cite{Geiger_2013_IJRR} \textit{val}-set for the car class. All methods are trained and evaluated on pseudo LiDAR point clouds generated by graphically corrected depth maps using \mbox{Pseudo-Lidar++~\cite{Yurong_2020_ICLR}}. The default beam selection is set at \{5,7,9,11\}. Our proposed method is trained using a reward function based on PointPillars~\cite{Lang_2019_CVPR}. The resulting beam configuration is given as \{7,8,9,10\} starting from a random initialization.}
    \label{tab:results_3dod}
    \tabcolsep=0.11cm
    \begin{tabular}{| l | c c c | c c c | c c c |} 
        \hline
          & \multicolumn{3}{c|}{Equidistant \{5, 7, 9, 11\}}  & \multicolumn{3}{c|}{\textepsilon-GS \{7, 8, 9, 10\}} & \multicolumn{3}{c|}{Difference} \\ 
        Method & Easy & Moderate & Hard & Easy & Moderate & Hard   & Easy & Moderate & Hard\\
        \hline
        PointPillars~\cite{Lang_2019_CVPR}  & 77.06          & 63.76          & 56.88  & \textbf{77.29} & \textbf{64.99} & \textbf{57.49} & \color{forestgreen}{+0.23} & \color{forestgreen}{+1.23} & \color{forestgreen}{+0.61}\\
    
        PointRCNN~\cite{Shi_2019_CVPR}  & \textbf{77.90} & 64.01          & 55.76     & 77.79          & \textbf{65.30} & \textbf{57.95} & \color{firebrick}{-0.11} & \color{forestgreen}{+1.29} & \color{forestgreen}{+2.19}\\
    
        SECOND~\cite{Yan_2018_Sensors} & 80.32 & 67.23 & 61.53 & \textbf{81.53} & \textbf{68.43} & \textbf{62.44} & \color{forestgreen}{+1.21} & \color{forestgreen}{+1.20} & \color{forestgreen}{+0.91}\\
    
        Part-\textit{A}\textsuperscript{2}~\cite{Shi_2021_TPAMI} & \textbf{86.19} & 69.12 & 65.52 & 85.91 & \textbf{71.82} & \textbf{66.99} & \color{firebrick}{-0.28} & \color{forestgreen}{+2.70} & \color{forestgreen}{+1.47} \\
        
        Voxel R-CNN~\cite{Deng_2021_AAAI} & 86.82 & 72.80 & 67.09 & \textbf{87.43} & \textbf{73.09} & \textbf{67.91} & \color{forestgreen}{+0.61} & \color{forestgreen}{+0.29} & \color{forestgreen}{+0.82} \\
        \hline
    \end{tabular}
    \vspace*{-.3cm}
\end{table*}

\section{Experiments}
\label{sec:experiments}

\subsection{Implementation Details of \texorpdfstring{$\varepsilon$}{e}-Greedy Search}
\label{sec:implementation_details}

To predict the value, we leverage a small fully connected network with two hidden layers with 128 and 64 nodes, respectively, using the ReLU activation function. A final layer outputs the predicted value signal passed through a sigmoid function. After each re-initialization, i.e., whenever the training set has been updated, we re-train the network for 10 epochs using the Adam optimizer guided by the mean squared error (MSE) loss.

In our implementation, the number of dimensions of the input feature $\mathbf{f}$ is $h=39$. To assign semantic labels to every point, we perform semantic segmentation on the temporally closest RGB images of each scan and then project the LiDAR point cloud into image space. Inspired by~\cite{Bruls_2019_ITSC}, we use a publicly available version\footnoteurl{https://github.com/tensorflow/models/tree/master/research/deeplab}{github.com/tensorflow/models/tree/master/\newline\indent research/deeplab} of DeepLab~\cite{Chen_2018_ECCV} with the Xception~\cite{Chollet_2017_CVPR} backbone, which has been trained on the Cityscapes dataset~\cite{Cordts_2016_CVPR}. If a point is projected onto multiple images with disagreeing labels, one of the candidates is picked at random. To reduce the feature space, and hence the required network size, we combine them into the following five superclasses based on their similar semantic meanings: road, building, vegetation, dynamic, and other objects. We demonstrate the effect on the network's reward prediction of each feature by separately adding them and report the prediction error in \reftab{tab:ablation_features_localization}. The mean absolute error (MAE) is computed by taking 100 and 200 random samples for object detection and localization, respectively, and using the remaining samples with ground truth (40 and 7,406) as a testing set. The error is averaged over 10 repetitions using different random seeds. As can be seen, combining all of our proposed features yields good performance across both tasks and clearly outperforms a naive encoding using only the beam IDs. As described in \refsec{sec:method_object_detection}, we use Pseudo-Lidar++~\cite{Yurong_2020_ICLR} for the object detection task. Since the method leverages a 4-beam LiDAR, we search for 4-beam configurations as well. Thus, we set $k=4$ for both tasks to enable a one-to-one comparison. For the action, we set the maximum step size $m=2$  balancing between rapid exploration of the search space and exploiting the more accurate predictions of the agent in the current area. For 3D object detection and localization, we set $K=40$\footnote{We only consider 40 out of the 64 available beams due to technical \newline\indent reasons, detailed in \refsec{sec:experiments_object_detection}.} and $K=32$, respectively. This is because the KITTI Dataset~\cite{Geiger_2013_IJRR} is used for object detection and the Oxford RobotCar Dataset~\cite{Barnes_2020_ICRA} is used for localization.

\subsection{3D Object Detection with Sparse Beam Selection}
\label{sec:experiments_object_detection}

For 3D object detection, we conduct experiments on the KITTI Dataset~\cite{Geiger_2013_IJRR}. We generate per image depth maps by using the model\footnoteurl{https://github.com/mileyan/Pseudo_Lidar_V2}{github.com/mileyan/Pseudo\_Lidar\_V2} from~\cite{Yurong_2020_ICLR} that has been trained on KITTI. After the depth-corrected pseudo LiDAR point clouds are generated, we resample 64 lines to retain sparsity.

When assigning beam IDs to the 64 lines, to generate a uniform distribution of points across all beams, we cluster the elevation angles into equidistant bin lengths starting from beam 64, with the bin length based on the LiDAR sensor's specification. This results in an out-of-distribution high point count for the initial beam 0 due to manufacturing tolerances, which therefore is discarded as a possible selection. Furthermore, beams above beam 40 are also discarded as they fall outside of the camera's field of view, resulting in the final set of $[1,40]$ $\subset \mathbb{N}$. As it is still infeasible to fully explore this set of 91,390 unique configurations, we leverage the proposed \textepsilon -GS to effectively navigate within this search space.

Due to its training speed and efficiency, we use a baseline PointPillars~\cite{Lang_2019_CVPR} to train on the pseudo LiDAR point clouds for 40 epochs with a learning rate of 0.003 using the Adam optimizer and a one cycle scheduler with a learning rate decay of 0.1 at epoch 35. The reward function in \refeqn{eq:reward_obj} for the RL-L2O algorithm is thus given by the 3D mAP performance on the \textit{val}-set for the moderate difficulty car class.

A single iteration of the search algorithm takes approximately 3.5h (1.5h pre-processing, 2h training) using 4 Nvidia GeForce GTX TITAN X (12GB) GPUs and an Intel\textsuperscript{\textregistered} Xeon\textsuperscript{\textregistered} E5-2560 v4 CPU at 2.20GHz with 24 cores. The time required by the agent for re-training and evaluation of the actions is 5s corresponding to 0.04\% of an iteration.

The algorithm reaches a high performance beam configuration \{7,8,9,10\} after only 64 steps from random initialization (\{1,6,19,40\}). An illustration of the evolution of the beam selection can be seen in \reffig{fig:search_evolution_obj}. The algorithm converges slowly to a state where all beams lie near the horizon, as this is where most (moderate difficulty) car points lie. As seen qualitatively, this improves the detection performance of the model by reducing the error propagated from the initial stereo depth estimation step.

Once converged, we fix the beam selection (\{7,8,9,10\}) in order to fully retrain on PointPillars~\cite{Lang_2019_CVPR} for 80 epochs and evaluate the 3D object detection performance on the \textit{val}-set. The results can be seen in \reftab{tab:results_3dod}. The beam configuration resulting from \textepsilon -GS (\{7,8,9,10\}) outperforms the equidistant beam selection of~\cite{Yurong_2020_ICLR} (\{5,7,9,11\}) substantially for the moderate car difficulty (+1.23\% mAP). Furthermore, this resulting beam configuration also generalizes well to other difficulty classes, showing improvements of 0.23\% and 0.61\% for the easy and hard difficulty classes, respectively.

\begin{table}
    \centering
    \caption{Mean absolute error (MAE) of the agent's reward prediction. The first row refers to a beam ID encoding.}
    \label{tab:ablation_features_localization}
    \resizebox{\linewidth}{!}{
        \begin{tabular}{| c | c | c | c | c | c | c | c |}
        \hline
        $f_{pts}$ & $f_{dist}$ & $f_{std\_dist}$ & $f_{sem\_pts}^{c}$ & $f_{\varphi\_diff}$ & $f_{\varphi}$ & $\text{MAE}_{obj}$ & $\text{MAE}_{loc}$ \\
        \hline
               &        &        &        &        &        & 0.01716 & 0.166 \\
        \xmark &        &        &        &        &        & 0.01285 & 0.259 \\
        \xmark & \xmark &        &        &        &        & 0.00881 & 0.143 \\
        \xmark & \xmark & \xmark &        &        &        & 0.00876 & 0.143 \\
        \xmark & \xmark & \xmark & \xmark &        &        & 0.00877 & 0.126 \\
        \xmark & \xmark & \xmark & \xmark & \xmark &        & 0.00898 & 0.125 \\
        \xmark & \xmark & \xmark & \xmark & \xmark & \xmark & \textbf{0.00867} & \textbf{0.121} \\
        \hline
        \end{tabular}
    }
\end{table}

To demonstrate the robustness of the final model selection when employing RL-L2O for 3D object detection, we conduct further experiments using the converged beam selection with PointPillars~\cite{Lang_2019_CVPR} on multiple state-of-the-art 3D object detectors\footnoteurl{https://github.com/open-mmlab/OpenPCDet}{github.com/open-mmlab/OpenPCDet} and report the results in \reftab{tab:results_3dod}. As observed, the beam selection of RL-L2O allows for better precision across the board and not only for the initial model~\cite{Lang_2019_CVPR} but for all methods, showing up to 1.21\%, 2.70\%, and 2.19\% improvements for the easy, moderate, and hard difficulty car classes, respectively.

\subsection{Beam Selection for LiDAR-Based Localization}
\label{sec:experiments_lidar_based_localization}

\noindent \textbf{Map Generation:}
The generated 3D map of Oxford is built using the recording of January 14, 2019, at 14:15:12 GMT. At this day and time, there was only a little overcast and the cameras did not suffer from overexposure. Additionally, the overall GNSS signal reception was considerably better than in other recordings.

Unlike~\cite{Sattler_2018_CVPR}, we do not optimize multiple local submaps but create a single globally consistent map. To lower the complexity of the pose graph and to simplify the optimization process, we filter successive poses with a minimum traveled distance of 1m and remove areas the mapping vehicle has visited twice keeping only a few overlaps to enforce loop closure. The graph consists of optimizable 3D poses (vertices) and ICP-based transforms (edges) between consecutive poses as well as ten manually selected loop closures. After construction, the graph is optimized with the iterative Levenberg-Marquardt solver. To speed up the optimization process, we initialize the pose graph with 2D positions based on radar odometry. Convergence is reached within 100,000 iterations.

Finally, we accumulate the point clouds of the right Velodyne LiDAR associated with the graph's poses and remove all points on  potentially dynamic objects from the map. This is done by performing semantic segmentation on the images of all four cameras as described in \refsec{sec:implementation_details}. We discard all points attributed to the following classes: person, rider, car, truck, bus, train, motorcycle, and bicycle.

\begin{figure}[t]
    \centering
    \includegraphics[width=\linewidth]{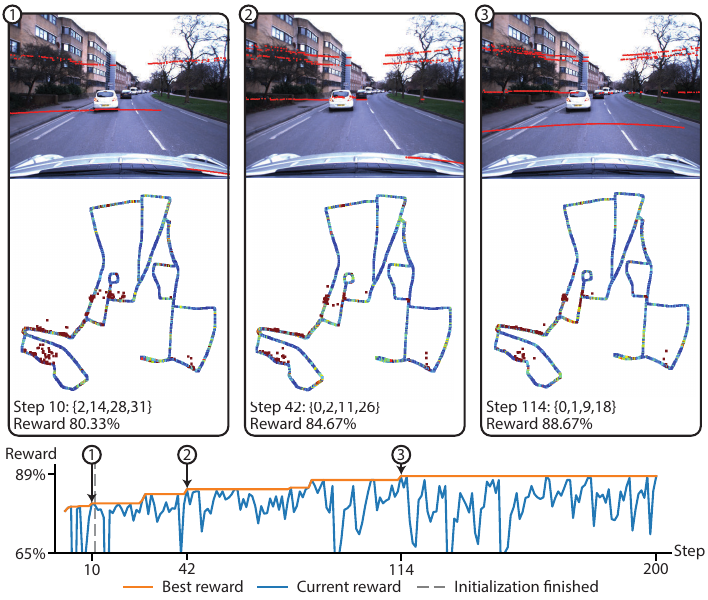}
    \caption{Evolution of the beam selection algorithm for LiDAR-based localization. (bottom) For the plot, please refer to the legend at the bottom. (top) Three cells are presented to visualize the best-performing beam configuration at different stages of the algorithm. The resulting LiDAR beams are shown on the camera image in red. The algorithm slowly converges to a state, where the beams are distributed such that constraints for all three spatial dimensions are incorporated while focusing on static objects. The lower part of each cell plots the lateral error of the entire route with blue and red denoting small and large errors, respectively.}
    \label{fig:search_evolution_loc}
\end{figure}

\begin{table}
    \centering
    \caption{Localization accuracy on the entire route for different matching thresholds. Our \textepsilon-GS is able to find a better beam configuration compared to a completely randomized search, i.e., $\varepsilon=1$. Both methods visited $T=200$ states. The equidistant baseline comprises beams evenly distributed across the range $[-6.67\degree, 6.67\degree]$. For reference, we also report the accuracy using the full 32-beam point cloud.}
    \label{tab:results_localization}
    \tabcolsep=0.11cm
    \resizebox{\linewidth}{!}{
        \begin{tabular}{| l | c | c c c |} 
        \hline
        & & \multicolumn{3}{c|}{Accuracy [\%]} \\
        Method & Beam IDs & .25m/2\degree & .50m/5\degree & 5.0m/10\degree \\
        \hline
        Full LiDAR & \{0, 1, \dots, 31\} & 89.87 & 96.52 & 96.64 \\
        \hline
        Equidistant & \{3, 6, 10, 13\} & 80.06 & 92.89 & 96.69 \\
        Random search & \{1, 5, 6, 22\} & 83.40 & 91.41 & 95.48 \\
        \textepsilon -GS & \{0, 1, 9, 18\} & \textbf{87.10} & \textbf{94.28} & \textbf{97.19} \\
        \hline
        \end{tabular}
    }
\end{table}

\begin{figure}
    \centering
    \includegraphics[width=\linewidth]{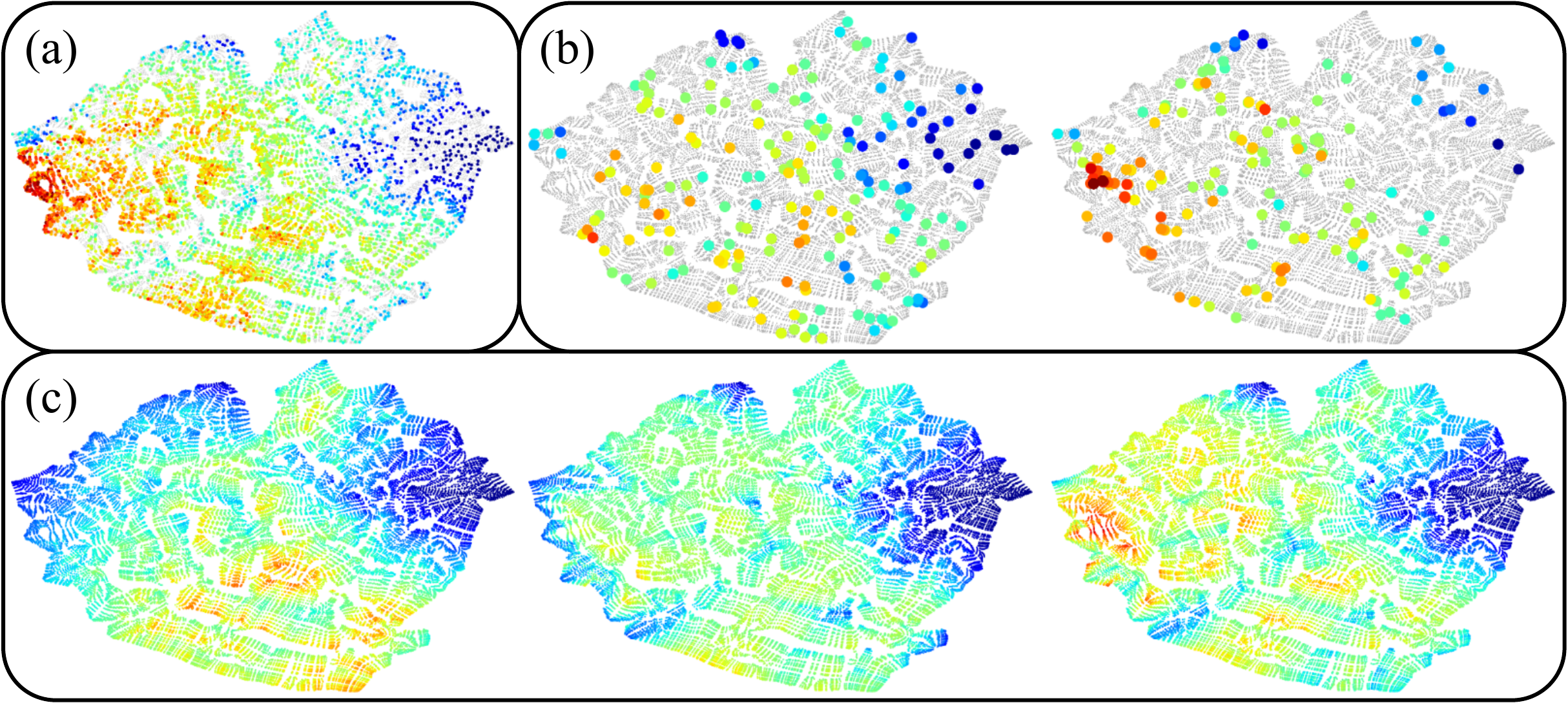}
    \caption{t-SNE representation of the search space of the localization task, where blue and red denote small and large values, respectively. (a) True values covering 20.6\% of the entire search space; (b) Comparison of random search (left) and \textepsilon-GS (right), which explores high-performing areas; (c) The values predicted by the agent while the search progresses for steps 10, 50, and 100 (left to right).}
    \label{fig:tsne_localization}
\end{figure}

\noindent \textbf{Beam Selection:}
We show the effectiveness of the proposed RL-L2O method as compared to random sampling, $\varepsilon=1$, using the same number of steps in both algorithms. To lower the run time, the search algorithms only have access to a subset of 100 poses, sampled randomly and then kept fixed for all experiments, from the recording of January 18, 2019, at 15:20:12 GMT. 
To emphasize more on the difficult accuracy thresholds, we set the weights, defined in \refeqn{eqn:reward_loc}, to $\lambda_1 = 3$, $\lambda_2 = 2$, and $\lambda_3 = 1$.

In particular, we run our RL-L2O method and random search ten times on the same poses while initializing with different random seeds in each repetition. In eight out of ten runs, RL-L2O has found an equal or more accurate beam configuration than random search when both search algorithms were executed for 200 steps, corresponding to $T$ in~\refalg{alg:eps_greedy}.
In our implementation, sampling 200 states takes approximately 13h, when executed on an Intel\textsuperscript{\textregistered} Core\textsuperscript{\texttrademark} i7-2600K CPU at 3.40GHz with 4 cores. The required time per step is similar for both methods as the only additional time is due to training the value function approximator and performing forward passes to evaluate possible actions, corresponding to an overhead of 5s. In the further evaluation, we refer to the results of the first run for both search algorithms. In \reftab{tab:results_localization}, we report the localization accuracy on the entire route, opposed to the 100 poses used during the search. The beam configuration found by our RL-L2O approach significantly outperforms the result of the random search. Compared to a naive baseline of equidistantly distributed beams in the range [-6.67\degree, +6.67\degree], the improvement, particularly for the hard threshold, is even more pronounced. For the easy threshold, our RL-L2O framework even surpasses the performance of using all 32 beams.

\reffig{fig:tsne_localization} shows a t-SNE~\cite{Maaten_2008_JMLR} representation of the 4D state space. As illustrated in (b-left), the data sampled by the random search is equally distributed among the space, whereas in (b-right), \textepsilon -GS explores the high-award areas. That is, after having visited the same number of states, \mbox{\textepsilon -GS} finds more beam configurations with a high localization accuracy than a purely randomized search, making it more sample efficient.

An illustration of the beam selection by \textepsilon -GS can be seen in \reffig{fig:search_evolution_loc}. Notable is that, in contrast to the results of 3D object detection, the found beams focus on static objects and are more widely distributed. This improves the localization accuracy since spanning over the entire z-direction adds a wide variety of spatial constraints to the underlying ICP algorithm. Additionally, we visualize in \reffig{fig:tsne_localization}~(c) how the reward prediction improves while the search progresses. As the agent gathers more samples, the predictions become more similar to the ground truth values, shown in (a).

\begin{figure} 
    \centering
    \captionsetup[subfigure]{justification=centering}
    \subfloat[Number of semantic points: road (violet), building (gray), vegetation (green), dynamic (red), and other (black).]{%
        \input{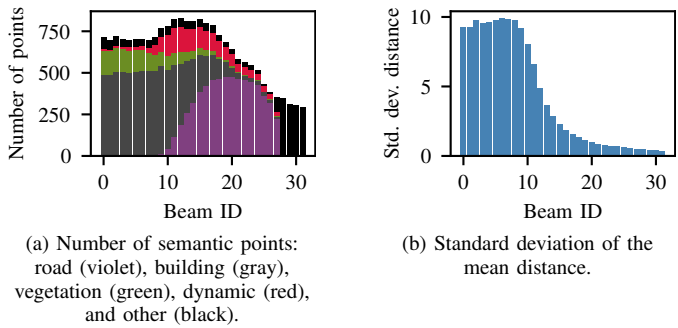}}
    \hfill
    \subfloat[Standard deviation of the mean distance.]{%
        \input{figures/beam_stats_std_distance.pgf}}
  \caption{Illustration of selected features used in the localization task. Beam ID~0 refers to the upward-facing beam with 10.67\degree, beam ID~31 refers to the downward-facing beam with -30.67\degree. The semantic labels in (a) are obtained from semantic segmentation on the synchronized RGB images.}
  \label{fig:feature_space_loc}
  \vspace{-.3cm}
\end{figure}

Finally, we illustrate the feature space of the localization task in \reffig{fig:feature_space_loc}. The resulting beam configuration \mbox{\{0, 1, 9, 18\}} concentrates on beams with a high variation in the depth measurements, an indicator for non-road points. As seen in~(a) and mentioned before, most beams focus on static classes such as buildings and vegetation. Note that the points of the beams 28 to 31 are projected outside the field of view of the cameras and hence are assigned the \textit{other} class.

%% file: figures/beam_stats_std_distance.pgf
\begingroup%
\makeatletter%
\begin{pgfpicture}%
\pgfpathrectangle{\pgfpointorigin}{\pgfqpoint{1.544535in}{1.203443in}}%
\pgfusepath{use as bounding box, clip}%
\begin{pgfscope}%
\pgfsetbuttcap%
\pgfsetmiterjoin%
\pgfsetlinewidth{0.000000pt}%
\definecolor{currentstroke}{rgb}{0.000000,0.000000,0.000000}%
\pgfsetstrokecolor{currentstroke}%
\pgfsetstrokeopacity{0.000000}%
\pgfsetdash{}{0pt}%
\pgfpathmoveto{\pgfqpoint{0.000000in}{0.000000in}}%
\pgfpathlineto{\pgfqpoint{1.544535in}{0.000000in}}%
\pgfpathlineto{\pgfqpoint{1.544535in}{1.203443in}}%
\pgfpathlineto{\pgfqpoint{0.000000in}{1.203443in}}%
\pgfpathclose%
\pgfusepath{}%
\end{pgfscope}%
\begin{pgfscope}%
\pgfsetbuttcap%
\pgfsetmiterjoin%
\pgfsetlinewidth{0.000000pt}%
\definecolor{currentstroke}{rgb}{0.000000,0.000000,0.000000}%
\pgfsetstrokecolor{currentstroke}%
\pgfsetstrokeopacity{0.000000}%
\pgfsetdash{}{0pt}%
\pgfpathmoveto{\pgfqpoint{0.369601in}{0.350309in}}%
\pgfpathlineto{\pgfqpoint{1.544535in}{0.350309in}}%
\pgfpathlineto{\pgfqpoint{1.544535in}{1.108183in}}%
\pgfpathlineto{\pgfqpoint{0.369601in}{1.108183in}}%
\pgfpathclose%
\pgfusepath{}%
\end{pgfscope}%
\begin{pgfscope}%
\pgfpathrectangle{\pgfqpoint{0.369601in}{0.350309in}}{\pgfqpoint{1.174935in}{0.757874in}}%
\pgfusepath{clip}%
\pgfsetbuttcap%
\pgfsetmiterjoin%
\definecolor{currentfill}{rgb}{0.274510,0.509804,0.705882}%
\pgfsetfillcolor{currentfill}%
\pgfsetlinewidth{0.000000pt}%
\definecolor{currentstroke}{rgb}{0.000000,0.000000,0.000000}%
\pgfsetstrokecolor{currentstroke}%
\pgfsetstrokeopacity{0.000000}%
\pgfsetdash{}{0pt}%
\pgfpathmoveto{\pgfqpoint{0.423007in}{0.350309in}}%
\pgfpathlineto{\pgfqpoint{0.449878in}{0.350309in}}%
\pgfpathlineto{\pgfqpoint{0.449878in}{1.024930in}}%
\pgfpathlineto{\pgfqpoint{0.423007in}{1.024930in}}%
\pgfpathclose%
\pgfusepath{fill}%
\end{pgfscope}%
\begin{pgfscope}%
\pgfpathrectangle{\pgfqpoint{0.369601in}{0.350309in}}{\pgfqpoint{1.174935in}{0.757874in}}%
\pgfusepath{clip}%
\pgfsetbuttcap%
\pgfsetmiterjoin%
\definecolor{currentfill}{rgb}{0.274510,0.509804,0.705882}%
\pgfsetfillcolor{currentfill}%
\pgfsetlinewidth{0.000000pt}%
\definecolor{currentstroke}{rgb}{0.000000,0.000000,0.000000}%
\pgfsetstrokecolor{currentstroke}%
\pgfsetstrokeopacity{0.000000}%
\pgfsetdash{}{0pt}%
\pgfpathmoveto{\pgfqpoint{0.456595in}{0.350309in}}%
\pgfpathlineto{\pgfqpoint{0.483466in}{0.350309in}}%
\pgfpathlineto{\pgfqpoint{0.483466in}{1.024065in}}%
\pgfpathlineto{\pgfqpoint{0.456595in}{1.024065in}}%
\pgfpathclose%
\pgfusepath{fill}%
\end{pgfscope}%
\begin{pgfscope}%
\pgfpathrectangle{\pgfqpoint{0.369601in}{0.350309in}}{\pgfqpoint{1.174935in}{0.757874in}}%
\pgfusepath{clip}%
\pgfsetbuttcap%
\pgfsetmiterjoin%
\definecolor{currentfill}{rgb}{0.274510,0.509804,0.705882}%
\pgfsetfillcolor{currentfill}%
\pgfsetlinewidth{0.000000pt}%
\definecolor{currentstroke}{rgb}{0.000000,0.000000,0.000000}%
\pgfsetstrokecolor{currentstroke}%
\pgfsetstrokeopacity{0.000000}%
\pgfsetdash{}{0pt}%
\pgfpathmoveto{\pgfqpoint{0.490184in}{0.350309in}}%
\pgfpathlineto{\pgfqpoint{0.517055in}{0.350309in}}%
\pgfpathlineto{\pgfqpoint{0.517055in}{1.058536in}}%
\pgfpathlineto{\pgfqpoint{0.490184in}{1.058536in}}%
\pgfpathclose%
\pgfusepath{fill}%
\end{pgfscope}%
\begin{pgfscope}%
\pgfpathrectangle{\pgfqpoint{0.369601in}{0.350309in}}{\pgfqpoint{1.174935in}{0.757874in}}%
\pgfusepath{clip}%
\pgfsetbuttcap%
\pgfsetmiterjoin%
\definecolor{currentfill}{rgb}{0.274510,0.509804,0.705882}%
\pgfsetfillcolor{currentfill}%
\pgfsetlinewidth{0.000000pt}%
\definecolor{currentstroke}{rgb}{0.000000,0.000000,0.000000}%
\pgfsetstrokecolor{currentstroke}%
\pgfsetstrokeopacity{0.000000}%
\pgfsetdash{}{0pt}%
\pgfpathmoveto{\pgfqpoint{0.523773in}{0.350309in}}%
\pgfpathlineto{\pgfqpoint{0.550644in}{0.350309in}}%
\pgfpathlineto{\pgfqpoint{0.550644in}{1.045088in}}%
\pgfpathlineto{\pgfqpoint{0.523773in}{1.045088in}}%
\pgfpathclose%
\pgfusepath{fill}%
\end{pgfscope}%
\begin{pgfscope}%
\pgfpathrectangle{\pgfqpoint{0.369601in}{0.350309in}}{\pgfqpoint{1.174935in}{0.757874in}}%
\pgfusepath{clip}%
\pgfsetbuttcap%
\pgfsetmiterjoin%
\definecolor{currentfill}{rgb}{0.274510,0.509804,0.705882}%
\pgfsetfillcolor{currentfill}%
\pgfsetlinewidth{0.000000pt}%
\definecolor{currentstroke}{rgb}{0.000000,0.000000,0.000000}%
\pgfsetstrokecolor{currentstroke}%
\pgfsetstrokeopacity{0.000000}%
\pgfsetdash{}{0pt}%
\pgfpathmoveto{\pgfqpoint{0.557362in}{0.350309in}}%
\pgfpathlineto{\pgfqpoint{0.584233in}{0.350309in}}%
\pgfpathlineto{\pgfqpoint{0.584233in}{1.051616in}}%
\pgfpathlineto{\pgfqpoint{0.557362in}{1.051616in}}%
\pgfpathclose%
\pgfusepath{fill}%
\end{pgfscope}%
\begin{pgfscope}%
\pgfpathrectangle{\pgfqpoint{0.369601in}{0.350309in}}{\pgfqpoint{1.174935in}{0.757874in}}%
\pgfusepath{clip}%
\pgfsetbuttcap%
\pgfsetmiterjoin%
\definecolor{currentfill}{rgb}{0.274510,0.509804,0.705882}%
\pgfsetfillcolor{currentfill}%
\pgfsetlinewidth{0.000000pt}%
\definecolor{currentstroke}{rgb}{0.000000,0.000000,0.000000}%
\pgfsetstrokecolor{currentstroke}%
\pgfsetstrokeopacity{0.000000}%
\pgfsetdash{}{0pt}%
\pgfpathmoveto{\pgfqpoint{0.590950in}{0.350309in}}%
\pgfpathlineto{\pgfqpoint{0.617821in}{0.350309in}}%
\pgfpathlineto{\pgfqpoint{0.617821in}{1.060637in}}%
\pgfpathlineto{\pgfqpoint{0.590950in}{1.060637in}}%
\pgfpathclose%
\pgfusepath{fill}%
\end{pgfscope}%
\begin{pgfscope}%
\pgfpathrectangle{\pgfqpoint{0.369601in}{0.350309in}}{\pgfqpoint{1.174935in}{0.757874in}}%
\pgfusepath{clip}%
\pgfsetbuttcap%
\pgfsetmiterjoin%
\definecolor{currentfill}{rgb}{0.274510,0.509804,0.705882}%
\pgfsetfillcolor{currentfill}%
\pgfsetlinewidth{0.000000pt}%
\definecolor{currentstroke}{rgb}{0.000000,0.000000,0.000000}%
\pgfsetstrokecolor{currentstroke}%
\pgfsetstrokeopacity{0.000000}%
\pgfsetdash{}{0pt}%
\pgfpathmoveto{\pgfqpoint{0.624539in}{0.350309in}}%
\pgfpathlineto{\pgfqpoint{0.651410in}{0.350309in}}%
\pgfpathlineto{\pgfqpoint{0.651410in}{1.072094in}}%
\pgfpathlineto{\pgfqpoint{0.624539in}{1.072094in}}%
\pgfpathclose%
\pgfusepath{fill}%
\end{pgfscope}%
\begin{pgfscope}%
\pgfpathrectangle{\pgfqpoint{0.369601in}{0.350309in}}{\pgfqpoint{1.174935in}{0.757874in}}%
\pgfusepath{clip}%
\pgfsetbuttcap%
\pgfsetmiterjoin%
\definecolor{currentfill}{rgb}{0.274510,0.509804,0.705882}%
\pgfsetfillcolor{currentfill}%
\pgfsetlinewidth{0.000000pt}%
\definecolor{currentstroke}{rgb}{0.000000,0.000000,0.000000}%
\pgfsetstrokecolor{currentstroke}%
\pgfsetstrokeopacity{0.000000}%
\pgfsetdash{}{0pt}%
\pgfpathmoveto{\pgfqpoint{0.658128in}{0.350309in}}%
\pgfpathlineto{\pgfqpoint{0.684999in}{0.350309in}}%
\pgfpathlineto{\pgfqpoint{0.684999in}{1.063565in}}%
\pgfpathlineto{\pgfqpoint{0.658128in}{1.063565in}}%
\pgfpathclose%
\pgfusepath{fill}%
\end{pgfscope}%
\begin{pgfscope}%
\pgfpathrectangle{\pgfqpoint{0.369601in}{0.350309in}}{\pgfqpoint{1.174935in}{0.757874in}}%
\pgfusepath{clip}%
\pgfsetbuttcap%
\pgfsetmiterjoin%
\definecolor{currentfill}{rgb}{0.274510,0.509804,0.705882}%
\pgfsetfillcolor{currentfill}%
\pgfsetlinewidth{0.000000pt}%
\definecolor{currentstroke}{rgb}{0.000000,0.000000,0.000000}%
\pgfsetstrokecolor{currentstroke}%
\pgfsetstrokeopacity{0.000000}%
\pgfsetdash{}{0pt}%
\pgfpathmoveto{\pgfqpoint{0.691717in}{0.350309in}}%
\pgfpathlineto{\pgfqpoint{0.718588in}{0.350309in}}%
\pgfpathlineto{\pgfqpoint{0.718588in}{1.061995in}}%
\pgfpathlineto{\pgfqpoint{0.691717in}{1.061995in}}%
\pgfpathclose%
\pgfusepath{fill}%
\end{pgfscope}%
\begin{pgfscope}%
\pgfpathrectangle{\pgfqpoint{0.369601in}{0.350309in}}{\pgfqpoint{1.174935in}{0.757874in}}%
\pgfusepath{clip}%
\pgfsetbuttcap%
\pgfsetmiterjoin%
\definecolor{currentfill}{rgb}{0.274510,0.509804,0.705882}%
\pgfsetfillcolor{currentfill}%
\pgfsetlinewidth{0.000000pt}%
\definecolor{currentstroke}{rgb}{0.000000,0.000000,0.000000}%
\pgfsetstrokecolor{currentstroke}%
\pgfsetstrokeopacity{0.000000}%
\pgfsetdash{}{0pt}%
\pgfpathmoveto{\pgfqpoint{0.725305in}{0.350309in}}%
\pgfpathlineto{\pgfqpoint{0.752176in}{0.350309in}}%
\pgfpathlineto{\pgfqpoint{0.752176in}{1.019135in}}%
\pgfpathlineto{\pgfqpoint{0.725305in}{1.019135in}}%
\pgfpathclose%
\pgfusepath{fill}%
\end{pgfscope}%
\begin{pgfscope}%
\pgfpathrectangle{\pgfqpoint{0.369601in}{0.350309in}}{\pgfqpoint{1.174935in}{0.757874in}}%
\pgfusepath{clip}%
\pgfsetbuttcap%
\pgfsetmiterjoin%
\definecolor{currentfill}{rgb}{0.274510,0.509804,0.705882}%
\pgfsetfillcolor{currentfill}%
\pgfsetlinewidth{0.000000pt}%
\definecolor{currentstroke}{rgb}{0.000000,0.000000,0.000000}%
\pgfsetstrokecolor{currentstroke}%
\pgfsetstrokeopacity{0.000000}%
\pgfsetdash{}{0pt}%
\pgfpathmoveto{\pgfqpoint{0.758894in}{0.350309in}}%
\pgfpathlineto{\pgfqpoint{0.785765in}{0.350309in}}%
\pgfpathlineto{\pgfqpoint{0.785765in}{0.936405in}}%
\pgfpathlineto{\pgfqpoint{0.758894in}{0.936405in}}%
\pgfpathclose%
\pgfusepath{fill}%
\end{pgfscope}%
\begin{pgfscope}%
\pgfpathrectangle{\pgfqpoint{0.369601in}{0.350309in}}{\pgfqpoint{1.174935in}{0.757874in}}%
\pgfusepath{clip}%
\pgfsetbuttcap%
\pgfsetmiterjoin%
\definecolor{currentfill}{rgb}{0.274510,0.509804,0.705882}%
\pgfsetfillcolor{currentfill}%
\pgfsetlinewidth{0.000000pt}%
\definecolor{currentstroke}{rgb}{0.000000,0.000000,0.000000}%
\pgfsetstrokecolor{currentstroke}%
\pgfsetstrokeopacity{0.000000}%
\pgfsetdash{}{0pt}%
\pgfpathmoveto{\pgfqpoint{0.792483in}{0.350309in}}%
\pgfpathlineto{\pgfqpoint{0.819354in}{0.350309in}}%
\pgfpathlineto{\pgfqpoint{0.819354in}{0.830321in}}%
\pgfpathlineto{\pgfqpoint{0.792483in}{0.830321in}}%
\pgfpathclose%
\pgfusepath{fill}%
\end{pgfscope}%
\begin{pgfscope}%
\pgfpathrectangle{\pgfqpoint{0.369601in}{0.350309in}}{\pgfqpoint{1.174935in}{0.757874in}}%
\pgfusepath{clip}%
\pgfsetbuttcap%
\pgfsetmiterjoin%
\definecolor{currentfill}{rgb}{0.274510,0.509804,0.705882}%
\pgfsetfillcolor{currentfill}%
\pgfsetlinewidth{0.000000pt}%
\definecolor{currentstroke}{rgb}{0.000000,0.000000,0.000000}%
\pgfsetstrokecolor{currentstroke}%
\pgfsetstrokeopacity{0.000000}%
\pgfsetdash{}{0pt}%
\pgfpathmoveto{\pgfqpoint{0.826072in}{0.350309in}}%
\pgfpathlineto{\pgfqpoint{0.852943in}{0.350309in}}%
\pgfpathlineto{\pgfqpoint{0.852943in}{0.704494in}}%
\pgfpathlineto{\pgfqpoint{0.826072in}{0.704494in}}%
\pgfpathclose%
\pgfusepath{fill}%
\end{pgfscope}%
\begin{pgfscope}%
\pgfpathrectangle{\pgfqpoint{0.369601in}{0.350309in}}{\pgfqpoint{1.174935in}{0.757874in}}%
\pgfusepath{clip}%
\pgfsetbuttcap%
\pgfsetmiterjoin%
\definecolor{currentfill}{rgb}{0.274510,0.509804,0.705882}%
\pgfsetfillcolor{currentfill}%
\pgfsetlinewidth{0.000000pt}%
\definecolor{currentstroke}{rgb}{0.000000,0.000000,0.000000}%
\pgfsetstrokecolor{currentstroke}%
\pgfsetstrokeopacity{0.000000}%
\pgfsetdash{}{0pt}%
\pgfpathmoveto{\pgfqpoint{0.859660in}{0.350309in}}%
\pgfpathlineto{\pgfqpoint{0.886531in}{0.350309in}}%
\pgfpathlineto{\pgfqpoint{0.886531in}{0.613809in}}%
\pgfpathlineto{\pgfqpoint{0.859660in}{0.613809in}}%
\pgfpathclose%
\pgfusepath{fill}%
\end{pgfscope}%
\begin{pgfscope}%
\pgfpathrectangle{\pgfqpoint{0.369601in}{0.350309in}}{\pgfqpoint{1.174935in}{0.757874in}}%
\pgfusepath{clip}%
\pgfsetbuttcap%
\pgfsetmiterjoin%
\definecolor{currentfill}{rgb}{0.274510,0.509804,0.705882}%
\pgfsetfillcolor{currentfill}%
\pgfsetlinewidth{0.000000pt}%
\definecolor{currentstroke}{rgb}{0.000000,0.000000,0.000000}%
\pgfsetstrokecolor{currentstroke}%
\pgfsetstrokeopacity{0.000000}%
\pgfsetdash{}{0pt}%
\pgfpathmoveto{\pgfqpoint{0.893249in}{0.350309in}}%
\pgfpathlineto{\pgfqpoint{0.920120in}{0.350309in}}%
\pgfpathlineto{\pgfqpoint{0.920120in}{0.556487in}}%
\pgfpathlineto{\pgfqpoint{0.893249in}{0.556487in}}%
\pgfpathclose%
\pgfusepath{fill}%
\end{pgfscope}%
\begin{pgfscope}%
\pgfpathrectangle{\pgfqpoint{0.369601in}{0.350309in}}{\pgfqpoint{1.174935in}{0.757874in}}%
\pgfusepath{clip}%
\pgfsetbuttcap%
\pgfsetmiterjoin%
\definecolor{currentfill}{rgb}{0.274510,0.509804,0.705882}%
\pgfsetfillcolor{currentfill}%
\pgfsetlinewidth{0.000000pt}%
\definecolor{currentstroke}{rgb}{0.000000,0.000000,0.000000}%
\pgfsetstrokecolor{currentstroke}%
\pgfsetstrokeopacity{0.000000}%
\pgfsetdash{}{0pt}%
\pgfpathmoveto{\pgfqpoint{0.926838in}{0.350309in}}%
\pgfpathlineto{\pgfqpoint{0.953709in}{0.350309in}}%
\pgfpathlineto{\pgfqpoint{0.953709in}{0.513578in}}%
\pgfpathlineto{\pgfqpoint{0.926838in}{0.513578in}}%
\pgfpathclose%
\pgfusepath{fill}%
\end{pgfscope}%
\begin{pgfscope}%
\pgfpathrectangle{\pgfqpoint{0.369601in}{0.350309in}}{\pgfqpoint{1.174935in}{0.757874in}}%
\pgfusepath{clip}%
\pgfsetbuttcap%
\pgfsetmiterjoin%
\definecolor{currentfill}{rgb}{0.274510,0.509804,0.705882}%
\pgfsetfillcolor{currentfill}%
\pgfsetlinewidth{0.000000pt}%
\definecolor{currentstroke}{rgb}{0.000000,0.000000,0.000000}%
\pgfsetstrokecolor{currentstroke}%
\pgfsetstrokeopacity{0.000000}%
\pgfsetdash{}{0pt}%
\pgfpathmoveto{\pgfqpoint{0.960427in}{0.350309in}}%
\pgfpathlineto{\pgfqpoint{0.987298in}{0.350309in}}%
\pgfpathlineto{\pgfqpoint{0.987298in}{0.483198in}}%
\pgfpathlineto{\pgfqpoint{0.960427in}{0.483198in}}%
\pgfpathclose%
\pgfusepath{fill}%
\end{pgfscope}%
\begin{pgfscope}%
\pgfpathrectangle{\pgfqpoint{0.369601in}{0.350309in}}{\pgfqpoint{1.174935in}{0.757874in}}%
\pgfusepath{clip}%
\pgfsetbuttcap%
\pgfsetmiterjoin%
\definecolor{currentfill}{rgb}{0.274510,0.509804,0.705882}%
\pgfsetfillcolor{currentfill}%
\pgfsetlinewidth{0.000000pt}%
\definecolor{currentstroke}{rgb}{0.000000,0.000000,0.000000}%
\pgfsetstrokecolor{currentstroke}%
\pgfsetstrokeopacity{0.000000}%
\pgfsetdash{}{0pt}%
\pgfpathmoveto{\pgfqpoint{0.994016in}{0.350309in}}%
\pgfpathlineto{\pgfqpoint{1.020887in}{0.350309in}}%
\pgfpathlineto{\pgfqpoint{1.020887in}{0.463759in}}%
\pgfpathlineto{\pgfqpoint{0.994016in}{0.463759in}}%
\pgfpathclose%
\pgfusepath{fill}%
\end{pgfscope}%
\begin{pgfscope}%
\pgfpathrectangle{\pgfqpoint{0.369601in}{0.350309in}}{\pgfqpoint{1.174935in}{0.757874in}}%
\pgfusepath{clip}%
\pgfsetbuttcap%
\pgfsetmiterjoin%
\definecolor{currentfill}{rgb}{0.274510,0.509804,0.705882}%
\pgfsetfillcolor{currentfill}%
\pgfsetlinewidth{0.000000pt}%
\definecolor{currentstroke}{rgb}{0.000000,0.000000,0.000000}%
\pgfsetstrokecolor{currentstroke}%
\pgfsetstrokeopacity{0.000000}%
\pgfsetdash{}{0pt}%
\pgfpathmoveto{\pgfqpoint{1.027604in}{0.350309in}}%
\pgfpathlineto{\pgfqpoint{1.054475in}{0.350309in}}%
\pgfpathlineto{\pgfqpoint{1.054475in}{0.448573in}}%
\pgfpathlineto{\pgfqpoint{1.027604in}{0.448573in}}%
\pgfpathclose%
\pgfusepath{fill}%
\end{pgfscope}%
\begin{pgfscope}%
\pgfpathrectangle{\pgfqpoint{0.369601in}{0.350309in}}{\pgfqpoint{1.174935in}{0.757874in}}%
\pgfusepath{clip}%
\pgfsetbuttcap%
\pgfsetmiterjoin%
\definecolor{currentfill}{rgb}{0.274510,0.509804,0.705882}%
\pgfsetfillcolor{currentfill}%
\pgfsetlinewidth{0.000000pt}%
\definecolor{currentstroke}{rgb}{0.000000,0.000000,0.000000}%
\pgfsetstrokecolor{currentstroke}%
\pgfsetstrokeopacity{0.000000}%
\pgfsetdash{}{0pt}%
\pgfpathmoveto{\pgfqpoint{1.061193in}{0.350309in}}%
\pgfpathlineto{\pgfqpoint{1.088064in}{0.350309in}}%
\pgfpathlineto{\pgfqpoint{1.088064in}{0.433002in}}%
\pgfpathlineto{\pgfqpoint{1.061193in}{0.433002in}}%
\pgfpathclose%
\pgfusepath{fill}%
\end{pgfscope}%
\begin{pgfscope}%
\pgfpathrectangle{\pgfqpoint{0.369601in}{0.350309in}}{\pgfqpoint{1.174935in}{0.757874in}}%
\pgfusepath{clip}%
\pgfsetbuttcap%
\pgfsetmiterjoin%
\definecolor{currentfill}{rgb}{0.274510,0.509804,0.705882}%
\pgfsetfillcolor{currentfill}%
\pgfsetlinewidth{0.000000pt}%
\definecolor{currentstroke}{rgb}{0.000000,0.000000,0.000000}%
\pgfsetstrokecolor{currentstroke}%
\pgfsetstrokeopacity{0.000000}%
\pgfsetdash{}{0pt}%
\pgfpathmoveto{\pgfqpoint{1.094782in}{0.350309in}}%
\pgfpathlineto{\pgfqpoint{1.121653in}{0.350309in}}%
\pgfpathlineto{\pgfqpoint{1.121653in}{0.422918in}}%
\pgfpathlineto{\pgfqpoint{1.094782in}{0.422918in}}%
\pgfpathclose%
\pgfusepath{fill}%
\end{pgfscope}%
\begin{pgfscope}%
\pgfpathrectangle{\pgfqpoint{0.369601in}{0.350309in}}{\pgfqpoint{1.174935in}{0.757874in}}%
\pgfusepath{clip}%
\pgfsetbuttcap%
\pgfsetmiterjoin%
\definecolor{currentfill}{rgb}{0.274510,0.509804,0.705882}%
\pgfsetfillcolor{currentfill}%
\pgfsetlinewidth{0.000000pt}%
\definecolor{currentstroke}{rgb}{0.000000,0.000000,0.000000}%
\pgfsetstrokecolor{currentstroke}%
\pgfsetstrokeopacity{0.000000}%
\pgfsetdash{}{0pt}%
\pgfpathmoveto{\pgfqpoint{1.128371in}{0.350309in}}%
\pgfpathlineto{\pgfqpoint{1.155242in}{0.350309in}}%
\pgfpathlineto{\pgfqpoint{1.155242in}{0.411414in}}%
\pgfpathlineto{\pgfqpoint{1.128371in}{0.411414in}}%
\pgfpathclose%
\pgfusepath{fill}%
\end{pgfscope}%
\begin{pgfscope}%
\pgfpathrectangle{\pgfqpoint{0.369601in}{0.350309in}}{\pgfqpoint{1.174935in}{0.757874in}}%
\pgfusepath{clip}%
\pgfsetbuttcap%
\pgfsetmiterjoin%
\definecolor{currentfill}{rgb}{0.274510,0.509804,0.705882}%
\pgfsetfillcolor{currentfill}%
\pgfsetlinewidth{0.000000pt}%
\definecolor{currentstroke}{rgb}{0.000000,0.000000,0.000000}%
\pgfsetstrokecolor{currentstroke}%
\pgfsetstrokeopacity{0.000000}%
\pgfsetdash{}{0pt}%
\pgfpathmoveto{\pgfqpoint{1.161959in}{0.350309in}}%
\pgfpathlineto{\pgfqpoint{1.188830in}{0.350309in}}%
\pgfpathlineto{\pgfqpoint{1.188830in}{0.406783in}}%
\pgfpathlineto{\pgfqpoint{1.161959in}{0.406783in}}%
\pgfpathclose%
\pgfusepath{fill}%
\end{pgfscope}%
\begin{pgfscope}%
\pgfpathrectangle{\pgfqpoint{0.369601in}{0.350309in}}{\pgfqpoint{1.174935in}{0.757874in}}%
\pgfusepath{clip}%
\pgfsetbuttcap%
\pgfsetmiterjoin%
\definecolor{currentfill}{rgb}{0.274510,0.509804,0.705882}%
\pgfsetfillcolor{currentfill}%
\pgfsetlinewidth{0.000000pt}%
\definecolor{currentstroke}{rgb}{0.000000,0.000000,0.000000}%
\pgfsetstrokecolor{currentstroke}%
\pgfsetstrokeopacity{0.000000}%
\pgfsetdash{}{0pt}%
\pgfpathmoveto{\pgfqpoint{1.195548in}{0.350309in}}%
\pgfpathlineto{\pgfqpoint{1.222419in}{0.350309in}}%
\pgfpathlineto{\pgfqpoint{1.222419in}{0.402312in}}%
\pgfpathlineto{\pgfqpoint{1.195548in}{0.402312in}}%
\pgfpathclose%
\pgfusepath{fill}%
\end{pgfscope}%
\begin{pgfscope}%
\pgfpathrectangle{\pgfqpoint{0.369601in}{0.350309in}}{\pgfqpoint{1.174935in}{0.757874in}}%
\pgfusepath{clip}%
\pgfsetbuttcap%
\pgfsetmiterjoin%
\definecolor{currentfill}{rgb}{0.274510,0.509804,0.705882}%
\pgfsetfillcolor{currentfill}%
\pgfsetlinewidth{0.000000pt}%
\definecolor{currentstroke}{rgb}{0.000000,0.000000,0.000000}%
\pgfsetstrokecolor{currentstroke}%
\pgfsetstrokeopacity{0.000000}%
\pgfsetdash{}{0pt}%
\pgfpathmoveto{\pgfqpoint{1.229137in}{0.350309in}}%
\pgfpathlineto{\pgfqpoint{1.256008in}{0.350309in}}%
\pgfpathlineto{\pgfqpoint{1.256008in}{0.399629in}}%
\pgfpathlineto{\pgfqpoint{1.229137in}{0.399629in}}%
\pgfpathclose%
\pgfusepath{fill}%
\end{pgfscope}%
\begin{pgfscope}%
\pgfpathrectangle{\pgfqpoint{0.369601in}{0.350309in}}{\pgfqpoint{1.174935in}{0.757874in}}%
\pgfusepath{clip}%
\pgfsetbuttcap%
\pgfsetmiterjoin%
\definecolor{currentfill}{rgb}{0.274510,0.509804,0.705882}%
\pgfsetfillcolor{currentfill}%
\pgfsetlinewidth{0.000000pt}%
\definecolor{currentstroke}{rgb}{0.000000,0.000000,0.000000}%
\pgfsetstrokecolor{currentstroke}%
\pgfsetstrokeopacity{0.000000}%
\pgfsetdash{}{0pt}%
\pgfpathmoveto{\pgfqpoint{1.262726in}{0.350309in}}%
\pgfpathlineto{\pgfqpoint{1.289597in}{0.350309in}}%
\pgfpathlineto{\pgfqpoint{1.289597in}{0.393033in}}%
\pgfpathlineto{\pgfqpoint{1.262726in}{0.393033in}}%
\pgfpathclose%
\pgfusepath{fill}%
\end{pgfscope}%
\begin{pgfscope}%
\pgfpathrectangle{\pgfqpoint{0.369601in}{0.350309in}}{\pgfqpoint{1.174935in}{0.757874in}}%
\pgfusepath{clip}%
\pgfsetbuttcap%
\pgfsetmiterjoin%
\definecolor{currentfill}{rgb}{0.274510,0.509804,0.705882}%
\pgfsetfillcolor{currentfill}%
\pgfsetlinewidth{0.000000pt}%
\definecolor{currentstroke}{rgb}{0.000000,0.000000,0.000000}%
\pgfsetstrokecolor{currentstroke}%
\pgfsetstrokeopacity{0.000000}%
\pgfsetdash{}{0pt}%
\pgfpathmoveto{\pgfqpoint{1.296314in}{0.350309in}}%
\pgfpathlineto{\pgfqpoint{1.323185in}{0.350309in}}%
\pgfpathlineto{\pgfqpoint{1.323185in}{0.388021in}}%
\pgfpathlineto{\pgfqpoint{1.296314in}{0.388021in}}%
\pgfpathclose%
\pgfusepath{fill}%
\end{pgfscope}%
\begin{pgfscope}%
\pgfpathrectangle{\pgfqpoint{0.369601in}{0.350309in}}{\pgfqpoint{1.174935in}{0.757874in}}%
\pgfusepath{clip}%
\pgfsetbuttcap%
\pgfsetmiterjoin%
\definecolor{currentfill}{rgb}{0.274510,0.509804,0.705882}%
\pgfsetfillcolor{currentfill}%
\pgfsetlinewidth{0.000000pt}%
\definecolor{currentstroke}{rgb}{0.000000,0.000000,0.000000}%
\pgfsetstrokecolor{currentstroke}%
\pgfsetstrokeopacity{0.000000}%
\pgfsetdash{}{0pt}%
\pgfpathmoveto{\pgfqpoint{1.329903in}{0.350309in}}%
\pgfpathlineto{\pgfqpoint{1.356774in}{0.350309in}}%
\pgfpathlineto{\pgfqpoint{1.356774in}{0.385337in}}%
\pgfpathlineto{\pgfqpoint{1.329903in}{0.385337in}}%
\pgfpathclose%
\pgfusepath{fill}%
\end{pgfscope}%
\begin{pgfscope}%
\pgfpathrectangle{\pgfqpoint{0.369601in}{0.350309in}}{\pgfqpoint{1.174935in}{0.757874in}}%
\pgfusepath{clip}%
\pgfsetbuttcap%
\pgfsetmiterjoin%
\definecolor{currentfill}{rgb}{0.274510,0.509804,0.705882}%
\pgfsetfillcolor{currentfill}%
\pgfsetlinewidth{0.000000pt}%
\definecolor{currentstroke}{rgb}{0.000000,0.000000,0.000000}%
\pgfsetstrokecolor{currentstroke}%
\pgfsetstrokeopacity{0.000000}%
\pgfsetdash{}{0pt}%
\pgfpathmoveto{\pgfqpoint{1.363492in}{0.350309in}}%
\pgfpathlineto{\pgfqpoint{1.390363in}{0.350309in}}%
\pgfpathlineto{\pgfqpoint{1.390363in}{0.383054in}}%
\pgfpathlineto{\pgfqpoint{1.363492in}{0.383054in}}%
\pgfpathclose%
\pgfusepath{fill}%
\end{pgfscope}%
\begin{pgfscope}%
\pgfpathrectangle{\pgfqpoint{0.369601in}{0.350309in}}{\pgfqpoint{1.174935in}{0.757874in}}%
\pgfusepath{clip}%
\pgfsetbuttcap%
\pgfsetmiterjoin%
\definecolor{currentfill}{rgb}{0.274510,0.509804,0.705882}%
\pgfsetfillcolor{currentfill}%
\pgfsetlinewidth{0.000000pt}%
\definecolor{currentstroke}{rgb}{0.000000,0.000000,0.000000}%
\pgfsetstrokecolor{currentstroke}%
\pgfsetstrokeopacity{0.000000}%
\pgfsetdash{}{0pt}%
\pgfpathmoveto{\pgfqpoint{1.397081in}{0.350309in}}%
\pgfpathlineto{\pgfqpoint{1.423952in}{0.350309in}}%
\pgfpathlineto{\pgfqpoint{1.423952in}{0.379283in}}%
\pgfpathlineto{\pgfqpoint{1.397081in}{0.379283in}}%
\pgfpathclose%
\pgfusepath{fill}%
\end{pgfscope}%
\begin{pgfscope}%
\pgfpathrectangle{\pgfqpoint{0.369601in}{0.350309in}}{\pgfqpoint{1.174935in}{0.757874in}}%
\pgfusepath{clip}%
\pgfsetbuttcap%
\pgfsetmiterjoin%
\definecolor{currentfill}{rgb}{0.274510,0.509804,0.705882}%
\pgfsetfillcolor{currentfill}%
\pgfsetlinewidth{0.000000pt}%
\definecolor{currentstroke}{rgb}{0.000000,0.000000,0.000000}%
\pgfsetstrokecolor{currentstroke}%
\pgfsetstrokeopacity{0.000000}%
\pgfsetdash{}{0pt}%
\pgfpathmoveto{\pgfqpoint{1.430669in}{0.350309in}}%
\pgfpathlineto{\pgfqpoint{1.457540in}{0.350309in}}%
\pgfpathlineto{\pgfqpoint{1.457540in}{0.377391in}}%
\pgfpathlineto{\pgfqpoint{1.430669in}{0.377391in}}%
\pgfpathclose%
\pgfusepath{fill}%
\end{pgfscope}%
\begin{pgfscope}%
\pgfpathrectangle{\pgfqpoint{0.369601in}{0.350309in}}{\pgfqpoint{1.174935in}{0.757874in}}%
\pgfusepath{clip}%
\pgfsetbuttcap%
\pgfsetmiterjoin%
\definecolor{currentfill}{rgb}{0.274510,0.509804,0.705882}%
\pgfsetfillcolor{currentfill}%
\pgfsetlinewidth{0.000000pt}%
\definecolor{currentstroke}{rgb}{0.000000,0.000000,0.000000}%
\pgfsetstrokecolor{currentstroke}%
\pgfsetstrokeopacity{0.000000}%
\pgfsetdash{}{0pt}%
\pgfpathmoveto{\pgfqpoint{1.464258in}{0.350309in}}%
\pgfpathlineto{\pgfqpoint{1.491129in}{0.350309in}}%
\pgfpathlineto{\pgfqpoint{1.491129in}{0.375796in}}%
\pgfpathlineto{\pgfqpoint{1.464258in}{0.375796in}}%
\pgfpathclose%
\pgfusepath{fill}%
\end{pgfscope}%
\begin{pgfscope}%
\pgfsetbuttcap%
\pgfsetroundjoin%
\definecolor{currentfill}{rgb}{0.000000,0.000000,0.000000}%
\pgfsetfillcolor{currentfill}%
\pgfsetlinewidth{0.803000pt}%
\definecolor{currentstroke}{rgb}{0.000000,0.000000,0.000000}%
\pgfsetstrokecolor{currentstroke}%
\pgfsetdash{}{0pt}%
\pgfsys@defobject{currentmarker}{\pgfqpoint{0.000000in}{-0.048611in}}{\pgfqpoint{0.000000in}{0.000000in}}{%
\pgfpathmoveto{\pgfqpoint{0.000000in}{0.000000in}}%
\pgfpathlineto{\pgfqpoint{0.000000in}{-0.048611in}}%
\pgfusepath{stroke,fill}%
}%
\begin{pgfscope}%
\pgfsys@transformshift{0.436442in}{0.350309in}%
\pgfsys@useobject{currentmarker}{}%
\end{pgfscope}%
\end{pgfscope}%
\begin{pgfscope}%
\definecolor{textcolor}{rgb}{0.000000,0.000000,0.000000}%
\pgfsetstrokecolor{textcolor}%
\pgfsetfillcolor{textcolor}%
\pgftext[x=0.436442in,y=0.253087in,,top]{\color{textcolor}\rmfamily\fontsize{8.000000}{9.600000}\selectfont 0}%
\end{pgfscope}%
\begin{pgfscope}%
\pgfsetbuttcap%
\pgfsetroundjoin%
\definecolor{currentfill}{rgb}{0.000000,0.000000,0.000000}%
\pgfsetfillcolor{currentfill}%
\pgfsetlinewidth{0.803000pt}%
\definecolor{currentstroke}{rgb}{0.000000,0.000000,0.000000}%
\pgfsetstrokecolor{currentstroke}%
\pgfsetdash{}{0pt}%
\pgfsys@defobject{currentmarker}{\pgfqpoint{0.000000in}{-0.048611in}}{\pgfqpoint{0.000000in}{0.000000in}}{%
\pgfpathmoveto{\pgfqpoint{0.000000in}{0.000000in}}%
\pgfpathlineto{\pgfqpoint{0.000000in}{-0.048611in}}%
\pgfusepath{stroke,fill}%
}%
\begin{pgfscope}%
\pgfsys@transformshift{0.772330in}{0.350309in}%
\pgfsys@useobject{currentmarker}{}%
\end{pgfscope}%
\end{pgfscope}%
\begin{pgfscope}%
\definecolor{textcolor}{rgb}{0.000000,0.000000,0.000000}%
\pgfsetstrokecolor{textcolor}%
\pgfsetfillcolor{textcolor}%
\pgftext[x=0.772330in,y=0.253087in,,top]{\color{textcolor}\rmfamily\fontsize{8.000000}{9.600000}\selectfont 10}%
\end{pgfscope}%
\begin{pgfscope}%
\pgfsetbuttcap%
\pgfsetroundjoin%
\definecolor{currentfill}{rgb}{0.000000,0.000000,0.000000}%
\pgfsetfillcolor{currentfill}%
\pgfsetlinewidth{0.803000pt}%
\definecolor{currentstroke}{rgb}{0.000000,0.000000,0.000000}%
\pgfsetstrokecolor{currentstroke}%
\pgfsetdash{}{0pt}%
\pgfsys@defobject{currentmarker}{\pgfqpoint{0.000000in}{-0.048611in}}{\pgfqpoint{0.000000in}{0.000000in}}{%
\pgfpathmoveto{\pgfqpoint{0.000000in}{0.000000in}}%
\pgfpathlineto{\pgfqpoint{0.000000in}{-0.048611in}}%
\pgfusepath{stroke,fill}%
}%
\begin{pgfscope}%
\pgfsys@transformshift{1.108217in}{0.350309in}%
\pgfsys@useobject{currentmarker}{}%
\end{pgfscope}%
\end{pgfscope}%
\begin{pgfscope}%
\definecolor{textcolor}{rgb}{0.000000,0.000000,0.000000}%
\pgfsetstrokecolor{textcolor}%
\pgfsetfillcolor{textcolor}%
\pgftext[x=1.108217in,y=0.253087in,,top]{\color{textcolor}\rmfamily\fontsize{8.000000}{9.600000}\selectfont 20}%
\end{pgfscope}%
\begin{pgfscope}%
\pgfsetbuttcap%
\pgfsetroundjoin%
\definecolor{currentfill}{rgb}{0.000000,0.000000,0.000000}%
\pgfsetfillcolor{currentfill}%
\pgfsetlinewidth{0.803000pt}%
\definecolor{currentstroke}{rgb}{0.000000,0.000000,0.000000}%
\pgfsetstrokecolor{currentstroke}%
\pgfsetdash{}{0pt}%
\pgfsys@defobject{currentmarker}{\pgfqpoint{0.000000in}{-0.048611in}}{\pgfqpoint{0.000000in}{0.000000in}}{%
\pgfpathmoveto{\pgfqpoint{0.000000in}{0.000000in}}%
\pgfpathlineto{\pgfqpoint{0.000000in}{-0.048611in}}%
\pgfusepath{stroke,fill}%
}%
\begin{pgfscope}%
\pgfsys@transformshift{1.444105in}{0.350309in}%
\pgfsys@useobject{currentmarker}{}%
\end{pgfscope}%
\end{pgfscope}%
\begin{pgfscope}%
\definecolor{textcolor}{rgb}{0.000000,0.000000,0.000000}%
\pgfsetstrokecolor{textcolor}%
\pgfsetfillcolor{textcolor}%
\pgftext[x=1.444105in,y=0.253087in,,top]{\color{textcolor}\rmfamily\fontsize{8.000000}{9.600000}\selectfont 30}%
\end{pgfscope}%
\begin{pgfscope}%
\definecolor{textcolor}{rgb}{0.000000,0.000000,0.000000}%
\pgfsetstrokecolor{textcolor}%
\pgfsetfillcolor{textcolor}%
\pgftext[x=0.957068in,y=0.098766in,,top]{\color{textcolor}\rmfamily\fontsize{8.000000}{9.600000}\selectfont Beam ID}%
\end{pgfscope}%
\begin{pgfscope}%
\pgfsetbuttcap%
\pgfsetroundjoin%
\definecolor{currentfill}{rgb}{0.000000,0.000000,0.000000}%
\pgfsetfillcolor{currentfill}%
\pgfsetlinewidth{0.803000pt}%
\definecolor{currentstroke}{rgb}{0.000000,0.000000,0.000000}%
\pgfsetstrokecolor{currentstroke}%
\pgfsetdash{}{0pt}%
\pgfsys@defobject{currentmarker}{\pgfqpoint{-0.048611in}{0.000000in}}{\pgfqpoint{-0.000000in}{0.000000in}}{%
\pgfpathmoveto{\pgfqpoint{-0.000000in}{0.000000in}}%
\pgfpathlineto{\pgfqpoint{-0.048611in}{0.000000in}}%
\pgfusepath{stroke,fill}%
}%
\begin{pgfscope}%
\pgfsys@transformshift{0.369601in}{0.350309in}%
\pgfsys@useobject{currentmarker}{}%
\end{pgfscope}%
\end{pgfscope}%
\begin{pgfscope}%
\definecolor{textcolor}{rgb}{0.000000,0.000000,0.000000}%
\pgfsetstrokecolor{textcolor}%
\pgfsetfillcolor{textcolor}%
\pgftext[x=0.213350in, y=0.311729in, left, base]{\color{textcolor}\rmfamily\fontsize{8.000000}{9.600000}\selectfont 0}%
\end{pgfscope}%
\begin{pgfscope}%
\pgfsetbuttcap%
\pgfsetroundjoin%
\definecolor{currentfill}{rgb}{0.000000,0.000000,0.000000}%
\pgfsetfillcolor{currentfill}%
\pgfsetlinewidth{0.803000pt}%
\definecolor{currentstroke}{rgb}{0.000000,0.000000,0.000000}%
\pgfsetstrokecolor{currentstroke}%
\pgfsetdash{}{0pt}%
\pgfsys@defobject{currentmarker}{\pgfqpoint{-0.048611in}{0.000000in}}{\pgfqpoint{-0.000000in}{0.000000in}}{%
\pgfpathmoveto{\pgfqpoint{-0.000000in}{0.000000in}}%
\pgfpathlineto{\pgfqpoint{-0.048611in}{0.000000in}}%
\pgfusepath{stroke,fill}%
}%
\begin{pgfscope}%
\pgfsys@transformshift{0.369601in}{0.714312in}%
\pgfsys@useobject{currentmarker}{}%
\end{pgfscope}%
\end{pgfscope}%
\begin{pgfscope}%
\definecolor{textcolor}{rgb}{0.000000,0.000000,0.000000}%
\pgfsetstrokecolor{textcolor}%
\pgfsetfillcolor{textcolor}%
\pgftext[x=0.213350in, y=0.675731in, left, base]{\color{textcolor}\rmfamily\fontsize{8.000000}{9.600000}\selectfont 5}%
\end{pgfscope}%
\begin{pgfscope}%
\pgfsetbuttcap%
\pgfsetroundjoin%
\definecolor{currentfill}{rgb}{0.000000,0.000000,0.000000}%
\pgfsetfillcolor{currentfill}%
\pgfsetlinewidth{0.803000pt}%
\definecolor{currentstroke}{rgb}{0.000000,0.000000,0.000000}%
\pgfsetstrokecolor{currentstroke}%
\pgfsetdash{}{0pt}%
\pgfsys@defobject{currentmarker}{\pgfqpoint{-0.048611in}{0.000000in}}{\pgfqpoint{-0.000000in}{0.000000in}}{%
\pgfpathmoveto{\pgfqpoint{-0.000000in}{0.000000in}}%
\pgfpathlineto{\pgfqpoint{-0.048611in}{0.000000in}}%
\pgfusepath{stroke,fill}%
}%
\begin{pgfscope}%
\pgfsys@transformshift{0.369601in}{1.078314in}%
\pgfsys@useobject{currentmarker}{}%
\end{pgfscope}%
\end{pgfscope}%
\begin{pgfscope}%
\definecolor{textcolor}{rgb}{0.000000,0.000000,0.000000}%
\pgfsetstrokecolor{textcolor}%
\pgfsetfillcolor{textcolor}%
\pgftext[x=0.154321in, y=1.039734in, left, base]{\color{textcolor}\rmfamily\fontsize{8.000000}{9.600000}\selectfont 10}%
\end{pgfscope}%
\begin{pgfscope}%
\definecolor{textcolor}{rgb}{0.000000,0.000000,0.000000}%
\pgfsetstrokecolor{textcolor}%
\pgfsetfillcolor{textcolor}%
\pgftext[x=0.098766in,y=0.729246in,,bottom,rotate=90.000000]{\color{textcolor}\rmfamily\fontsize{8.000000}{9.600000}\selectfont Std. dev. distance}%
\end{pgfscope}%
\begin{pgfscope}%
\pgfsetrectcap%
\pgfsetmiterjoin%
\pgfsetlinewidth{0.803000pt}%
\definecolor{currentstroke}{rgb}{0.000000,0.000000,0.000000}%
\pgfsetstrokecolor{currentstroke}%
\pgfsetdash{}{0pt}%
\pgfpathmoveto{\pgfqpoint{0.369601in}{0.350309in}}%
\pgfpathlineto{\pgfqpoint{0.369601in}{1.108183in}}%
\pgfusepath{stroke}%
\end{pgfscope}%
\begin{pgfscope}%
\pgfsetrectcap%
\pgfsetmiterjoin%
\pgfsetlinewidth{0.803000pt}%
\definecolor{currentstroke}{rgb}{0.000000,0.000000,0.000000}%
\pgfsetstrokecolor{currentstroke}%
\pgfsetdash{}{0pt}%
\pgfpathmoveto{\pgfqpoint{1.544535in}{0.350309in}}%
\pgfpathlineto{\pgfqpoint{1.544535in}{1.108183in}}%
\pgfusepath{stroke}%
\end{pgfscope}%
\begin{pgfscope}%
\pgfsetrectcap%
\pgfsetmiterjoin%
\pgfsetlinewidth{0.803000pt}%
\definecolor{currentstroke}{rgb}{0.000000,0.000000,0.000000}%
\pgfsetstrokecolor{currentstroke}%
\pgfsetdash{}{0pt}%
\pgfpathmoveto{\pgfqpoint{0.369601in}{0.350309in}}%
\pgfpathlineto{\pgfqpoint{1.544535in}{0.350309in}}%
\pgfusepath{stroke}%
\end{pgfscope}%
\begin{pgfscope}%
\pgfsetrectcap%
\pgfsetmiterjoin%
\pgfsetlinewidth{0.803000pt}%
\definecolor{currentstroke}{rgb}{0.000000,0.000000,0.000000}%
\pgfsetstrokecolor{currentstroke}%
\pgfsetdash{}{0pt}%
\pgfpathmoveto{\pgfqpoint{0.369601in}{1.108183in}}%
\pgfpathlineto{\pgfqpoint{1.544535in}{1.108183in}}%
\pgfusepath{stroke}%
\end{pgfscope}%
\end{pgfpicture}%
\makeatother%
\endgroup%

%% file: sections/5_conclusion.tex
\section{Conclusion}
\label{sec:conclusion}

In this work, we introduce the problem of optimizing the position of beams of a low-resolution LiDAR while improving the performance for different LiDAR-based applications. As a possible solution, we presented a novel learning method based on reinforcement learning, which is end-to-end trainable and can be integrated with existing LiDAR-based methods as a simple drop-in module. Finally, we applied our proposed RL-L2O method to two core tasks of robotics, namely 3D object detection and localization. Extensive experiments demonstrated the efficacy of our approach for finding beam configurations that significantly improve the results of the respective baseline methods. We further showed that the retrieved beam configurations are indeed task-specific and are well aligned with human intuition. We expect that with the rise of reprogrammable LiDARs future work will also consider the case of dynamically adjusting the beam configuration based on the current scene and task. For instance, as the priors for LiDAR-based localization deviate between driving on a highway or in an urban environment, the beam positions should be adjusted accordingly.